\documentclass[journal,twoside,web]{ieeecolor}

\usepackage{balance}
\usepackage{tabularx}
\usepackage{framed,multirow}
\usepackage{tmi}
\usepackage{cite}
\usepackage{amsmath,amssymb,amsfonts}
\usepackage{algorithmic}
\usepackage{graphicx}
\usepackage{textcomp}
\usepackage{xcolor}
\usepackage{hyperref}
\usepackage{url}
\usepackage{xurl}

\makeatletter
\def\thickhline{%
	\noalign{\ifnum0=`}\fi\hrule \@height \thickarrayrulewidth \futurelet
	\reserved@a\@xthickhline}
\def\@xthickhline{\ifx\reserved@a\thickhline
	\vskip\doublerulesep
	\vskip-\thickarrayrulewidth
	\fi
	\ifnum0=`{\fi}}
\makeatother

\newlength{\thickarrayrulewidth}
\newcolumntype{L}{>{\raggedright\arraybackslash}X}

\newcolumntype{C}{>{\centering\arraybackslash}X}
\usepackage{url}
\usepackage{xcolor}

\usepackage{hyperref}

\usepackage{array}
\definecolor{newcolor}{rgb}{.8,.349,.1}

\def\BibTeX{{\rm B\kern-.05em{\sc i\kern-.025em b}\kern-.08em
    T\kern-.1667em\lower.7ex\hbox{E}\kern-.125emX}}
\begin{document}
\title{Bidirectional Semi-supervised Dual-branch CNN for Robust 3D Reconstruction of Stereo Endoscopic Images via Adaptive Cross and Parallel Supervisions}
\author{Hongkuan Shi, Zhiwei Wang, Ying Zhou, Dun Li, Xin Yang, \IEEEmembership{Member, IEEE}, Qiang Li, \IEEEmembership{Member, IEEE},
\thanks{This work was supported in part by National Natural Science Foundation of China (Grant No. 62202189), the National Key R\&D Program of China (Grant No. 2022YFE0200600), Fundamental Research Funds for the Central Universities (2021XXJS033), and research grants from Wuhan United Imaging Healthcare Surgical Technology Co., Ltd. \emph{(Hongkuan Shi and Zhiwei Wang are co-first authors.) (Corresponding author: Qiang Li.)}}
\thanks{Hongkuan Shi, Zhiwei Wang, Ying Zhou, Qiang Li are with Britton Chance Center for Biomedical Photonics, Wuhan National Laboratory for Optoelectronics, Huazhong University of Science and Technology, Wuhan, 430074, China. Dun Li is with United Imaging Surgical Healthcare Co., Ltd. 99 Gaokeyuan Rd., Wuhan, 430074, China. Xin Yang is with School of Electronic Information and Communications, Huazhong University of Science and Technology, Wuhan, 430074, China. (email:shihk@hust.edu.cn, zwwang@hust.edu.cn, zhouying17@hust.edu.cn, alex.li@united-imaging.com, xinyang2014@hust.edu.cn, liqiang8@hust.edu.cn)}
}

\maketitle

\begin{abstract}
Semi-supervised learning via teacher-student network can train a model effectively on a few labeled samples. It enables a student model to distill knowledge from the teacher’s predictions of extra unlabeled data. However, such knowledge flow is typically unidirectional, having the accuracy vulnerable to the quality of teacher model. In this paper, we seek to robust 3D reconstruction of stereo endoscopic images by proposing a novel fashion of \emph{bidirectional} learning between two learners, each of which can play both roles of teacher and student concurrently. Specifically, we introduce two self-supervisions, i.e., \emph{Adaptive Cross Supervision} (ACS) and \emph{Adaptive Parallel Supervision} (APS), to learn a dual-branch convolutional neural network. The two branches predict two different disparity probability distributions for the same position, and output their expectations as disparity values. The learned knowledge flows across branches along two directions: a cross direction (disparity guides distribution in ACS) and a parallel direction (disparity guides disparity in APS). Moreover, each branch also learns confidences to dynamically refine its provided supervisions. In ACS, the predicted disparity is softened into a unimodal distribution, and the lower the confidence, the smoother the distribution. In APS, the incorrect predictions are suppressed by lowering the weights of those with low confidence. With the adaptive bidirectional learning, the two branches enjoy well-tuned mutual supervisions, and eventually converge on a consistent and more accurate disparity estimation.
The experimental results on four public datasets demonstrate our superior accuracy over other state-of-the-arts with a relative decrease of averaged disparity error by at least 9.76\%.
\end{abstract}


\vspace{0.3cm}
\section{Introduction}
%\vspace{-0.15cm}
\IEEEPARstart{I}{n} robot assisted minimally invasive surgery (RA-MIS), 3D reconstruction of the stereo-endoscopic scenes is an enabling step for preoperative model registration, intra-operative surgical planning and so on \cite{stoyanov2008intra,taylor2016medical}. An accurate and dense stereo-endoscopic 3D reconstruction is important towards deployment of surgical guidance and visualization in RA-MIS\cite{stoyanov2008intra,maier2014comparative}, but extremely challenging because of the surgical scenes’ low-textured appearance, occlusions and especially scarce labels due to \emph{in-vivo} environments.

\begin{figure}[t]
	\includegraphics[width=\linewidth]{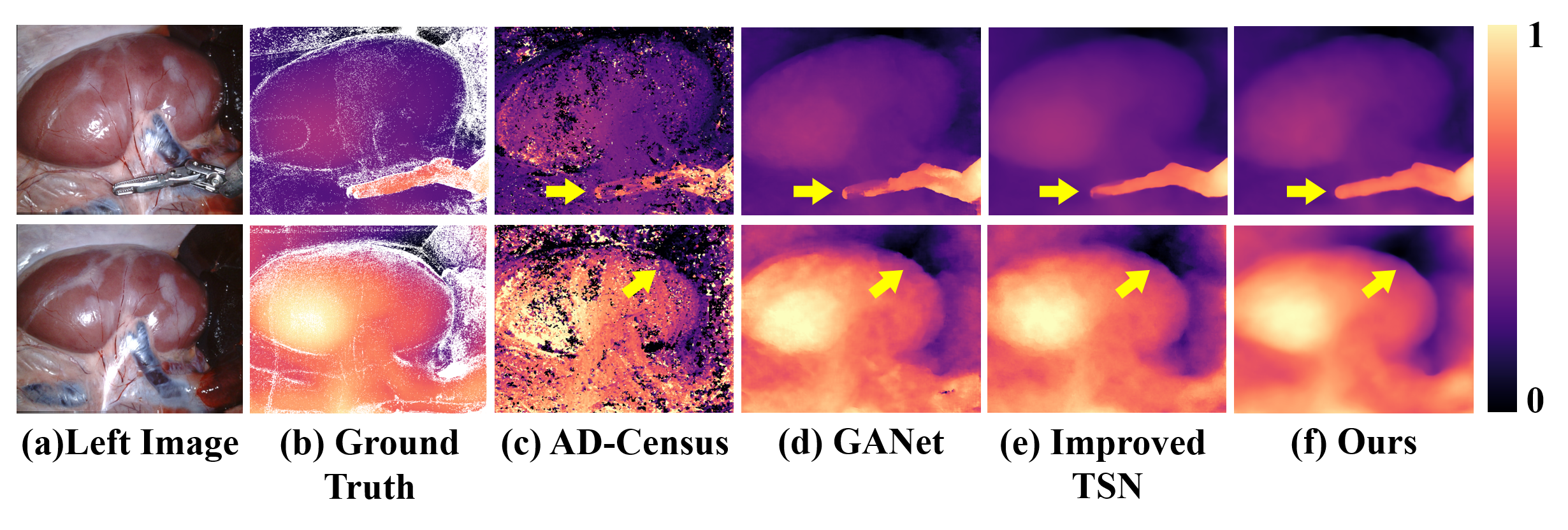}
	\caption{ Two examples of disparity estimation by different methods. (a) Left input image, (b) normalized ground truth disparity, (c-f) normalized disparity map predicted by traditional method AD-Census \cite{mei2011building}, fully-supervised method GAnet \cite{zhang2019ga}, Improved TSN \cite{shi2021semi}, and our method. The arrows indicate the differences in challenging regions of the disparity maps.}
	\vspace{-0.3cm}
	\label{fig:traditional}
\end{figure}

3D reconstruction of stereo images requires recovering a depth map by finding matching pixel locations along the horizontal epipolar line in a rectified left-right image pair. The depth is inversely proportional to the disparity, and the latter is defined as a pixel's shifting distance along the epipolar line from the left image to its corresponding position in the right image.
Traditional methods \cite{mei2011building,chang2013real,penza2016dense} typically rely on intensity similarities for matching pixels. Thus, low-textured surfaces in endoscopic scenes could make most traditional methods fail, yielding incorrect sparse disparity maps as evidenced in Fig. \ref{fig:traditional}~(c).

Recently, the methods based on Convolutional Neural Networks (CNN) have shown their impressive 3D reconstruction performance in both medical \cite{shi2021semi,rau2019implicit,allan2021stereo,mahmood2018deep,luo2019details,cui2019semi,li2018semi} and non-medical \cite{zhang2019ga,mayer2016large,godard2017unsupervised,ye2017self,huang2021self,smolyanskiy2018importance,laine2016temporal,tosi2019learning} domains. Depending on the label dependency, these methods can be categorized into \emph{fully-supervised}, \emph{self-supervised}, and \emph{semi-supervised} approaches. Fully-supervised methods \cite{mayer2016large,zhang2019ga} usually have an outstanding accuracy, but this comes with a price of laborious annotation of collected samples (i.e., the ground-truth (GT) disparity maps). For the stereo-endoscopic images, the GT map is hard or even infeasible to acquire, especially for some \emph{in-vivo} scenes. A CNN-based method trained on insufficient labeled samples usually suffers from an overfitting problem, which degrades its accuracy consequently. For example, Fig. \ref{fig:traditional}~(d) shows the results predicted by a recent fully-supervised method, i.e., GANet \cite{zhang2019ga}, trained on only 25 available labeled stereo endoscopic images.

The self-supervised methods \cite{godard2017unsupervised,ye2017self,huang2021self} learn stereo disparity estimation models with no labeled samples. They first estimate a disparity map aligned with the left image, and then reconstruct a synthesized left image by warping the right image along the epipolar line according to the estimated disparities. Therefore, the learning objective is to minimize pixel-wise reconstruction loss between the real and synthesized left images. 
However, such pixel-wise loss is unreliable because of the inconsistent brightness, which is caused by a strong non-Lambertian reflection on the organ surfaces. In addition to brightness differences, there may be surfaces with not enough texture/features to be able to match corresponding pixels, especially with little difference/parallax in camera views.
%However, the pixel-wise loss is unreliable in endoscopic scenarios. Inconsistent brightness caused by strong non-Lambertian reflections and surfaces with insufficient texture/features makes it difficult to find matching pixels, especially for the pixels with little difference or disparity in camera views.

The semi-supervised methods \cite{shi2021semi,smolyanskiy2018importance,luo2019details,laine2016temporal,cui2019semi,li2018semi,tosi2019learning} combine the strengths of both fully-supervised and self-supervised learnings. The supervised learning helps a more accurate disparity estimation by explicit guidance of GT maps, and the self-supervised learning enhances the generalization by distilling knowledge from the unlabeled samples. Among them, Teacher Student Network (TSN) based semi-supervised methods \cite{laine2016temporal,cui2019semi,li2018semi} have made great progress. In TSN, a teacher model is first trained in a supervised manner, and then predicts pseudo labels for unlabeled samples. A student model thus can be trained utilizing both labeled and unlabeled samples for further improving its accuracy. However, TSN often faces a dilemma when it comes to the medical data, e.g., endoscopic scenes, that is, the supervised teacher model hardly predicts highly confident pseudo labels due to few labeled data available for training, and teaches out a worse student model consequently.

In our previous work \cite{shi2021semi}, we have improved TSN to address the above issue by introducing a semi-supervised teacher model to predict mostly correct pseudo labels, and a confidence network to further suppress the unreliable predictions with low confidence. Therefore, a more accurate student model can be trained by use of more reliable pseudo labels. Despite our success, the improved TSN still has limitations, for example, the teacher model has to be separately trained beforehand, and the knowledge flow is unidirectional from the teacher to student. Thus, the teacher's accuracy becomes decisive, and its stagnant learning quality limits the possibility of a further improvement of the student’s accuracy.

In this paper, we break through the limitations, and propose a unified framework where the teacher-student and confidence networks can be trained jointly in a bidirectional semi-supervised fashion. As shown in the top of Fig. \ref{fig:two_branch}, the unified framework is essentially a dual-branch CNN. Each branch utilizes a disparity estimation network ($ DEnet $) and a confidence network ($ Confnet $) to predict three maps of disparity probability distribution, disparity value, and confidence. The value map is an expectation of the distribution map.

In the fully-supervised learning, each branch is trained separately on labeled samples, to predict their disparity value maps and corresponding confidence maps. In the self-supervised learning, \emph{Adaptive Cross Supervision} (ACS) and \emph{Adaptive Parallel supervision} (APS) are introduced to have the two branches mutually guide each other by taking the opposite’s predictions as supervisions. Specifically, ACS constrains each branch to predict a unimodal probability distribution with its peak aligned with the other’s disparity value, and meanwhile APS minimizes the L1 distance between the disparity values in the two branches. Moreover, the confidence adaptively controls each branch’s contribution to the learning of the other branch, and enhances the reliability of provided supervisions. That is, a lower confidence indicates a lower and wider peak of the unimodal distribution in ACS, and a less contribution of the supervision by a re-weighting strategy in APS, and vice versa.

In summary, our main contributions are listed:
\begin{enumerate}
	\item We surmount the limitations of our previous improved TSN \cite{shi2021semi}, and develop a novel semi-supervised dual-branch CNN for disparity estimation of stereo-endoscopic images. The two branches can mutually guide each other in a fashion of bidirectional learning, and eventually converge on a consistent and more accurate disparity estimation (see Fig. \ref{fig:traditional} (e)-(f)).
	%	 comparing with the semi-supervised TSN whose knowledge flow is unidirectional.
	\item We introduce ACS and APS as two kinds of the bidirectional supervisions, where the knowledge of each branch can be adaptively refined and flow along both cross and parallel directions to guide the learning of the other branch. The resulting well-tuned bidirectional supervisions maximize the efficacy of unlabeled data in the learning of our proposed dual-branch CNN.
	\item The extensive and comprehensive experimental results on four public datasets demonstrate the effectiveness of two proposed bidirectional supervisions, and a superior accuracy of our method over the fully-supervised and semi-supervised state-of-the-arts with a relative decrease of average disparity error by  12.94\% and 9.76\% at least, respectively. The source code is available\footnote{\url{https://github.com/HK-Shi/Bidirectional-SemiSupervised-Dual-branch-CNN}}.
	
\end{enumerate}

Also, this work differs from our conference work\cite{shi2021semi} from the following main aspects:
\begin{enumerate}
	\item Instead of separate optimization of the teacher $ DEnet $, student $ DEnet $ and confidence networks, this work develops a Dual-Branch CNN, where each branch contains a $ DEnet $ and a $ Confnet $, and plays both roles of teacher and student, concurrently. All networks of the two branches are optimized jointly in this work.
	\item Instead of unidirectional constraint of disparity value predictions from the teacher to student, this work introduces APS and ACS for bidirectional constraints of both disparity value and distribution predictions.
	\item In this work, we include three more datasets (i.e., SERV-CT, KITTI and ETH3D), eight more comparison state-of-the-art methods (i.e., four fully-supervised methods HSMNet~\cite{yang2019hierarchical}, CDN~\cite{garg2020wasserstein}, CFNet~\cite{shen2021cfnet} and ACFNet~\cite{zhang2020adaptive}, four semi-supervised methods Smolyanskiy \textit{et al.}~\cite{smolyanskiy2018importance}, SoftMT~\cite{xu2021end}, Tonioni \textit{et al.}~\cite{tonioni2019unsupervised} and Improved TSN~\cite{shi2021semi}) and more comprehensive ablation studies to verify the effectiveness of our method.
	
\end{enumerate}

%\vspace{-0.15cm}
\section{Related Work}
%\vspace{-0.15cm}

In this section, we review recent depth or disparity estimation methods related to fully-supervised, self-supervised and semi-supervised approaches.

\vspace{-0.15cm}
\subsection{Fully-supervised Depth/Disparity Estimation}
\label{SEC2.1}
The fully-supervised methods are characterized by a decent accuracy and large amounts of labeled samples for training. In natural scenes, the labeled data is relatively easy to acquire by filming using LiDAR and stereo camera simultaneously. For example, the commonly used outdoor datasets KITTI \cite{Geiger2012CVPR,Menze2015CVPR} provide over 29K training images with LiDAR GT maps. The fully-supervised methods thus focus on developing sophisticated and interpretable modules or techniques for an accurate depth/disparity estimation. 

Cost Volume (CV) \cite{mayer2016large,kendall2017end,chang2018pyramid,guo2019group,zhang2019ga} is one of the most successful techniques for a scene-invariant disparity estimation, and it constructs a disparity searching space by comparing the left and right images or features at different disparity levels. For example, DispNetC \cite{mayer2016large} first shifted right-ward the right feature map multiple times, and then computed inner product between each shifted right and the original left feature maps pixel-by-pixel to generate multiple scene-invariant feature correlation maps at different disparity levels. 
Therefore, the CV constructed by stacking all feature correlation maps is a 3D tensor of height, width and disparity.
Based on DispNet, GCNet\cite{kendall2017end} constructed a 4D CV by concatenating left-right feature maps at each disparity level, and used CNN to learn a more complex correlation function. 
With the learned correlation function, the CV is converted to the probability distribution. The disparity is computed as the expectation of the distribution.

Follow-up works\cite{zhang2020adaptive,garg2020wasserstein} consider that the CV models matching process but the network learning are driven by supervising the final disparity, which makes the network prone to overfitting since the CV is not constrained directly.
ACFNet\cite{zhang2020adaptive} generated unimodal probability distribution based on GT and estimated confidence, and proposed a stereo focal loss to constrain the CV.
CDN\cite{garg2020wasserstein} proposed a continuous disparity network to predict an offset for each pre-defined discrete disparity value, turning a categorical distribution to a continuous distribution. 
Despite the success of CV-based methods on natural scenes, GT disparity maps are hard to acquire for the stereo-endoscopic images. The resulting scarce labeled samples often lead the fully-supervised methods to overfitting, and degrade their prediction accuracy significantly.

To increase the number of labeled samples in medical domain, several methods \cite{mahmood2018deep,rau2019implicit} resorted to simulations based on virtual phantoms. For instance, Rau \textit{et al.} \cite{rau2019implicit} built a colon phantom using CT data, and generated a plenty of synthetic colon images and depth maps. After training a CNN model on the synthetic samples, a few real colon images were utilized to adapt the model onto the real scenarios.
However, it is hard to generate some properties, e.g., tissue deformation, specific to \emph{in-vivo} environment in the simulated data, therefore, there is a huge gap between the training and test scenes, and it is very difficult to bridge this gap by simply using a domain adaptation.

%\vspace{-0.15cm}
\subsection{Self-Supervised Depth/Disparity Estimation}
\label{SEC2.2}
Self-supervised methods \cite{godard2017unsupervised,ye2017self,huang2021self} constrained the disparity estimation via a proxy task of view reconstruction without using any GT depth/disparity map. The basic idea is that if the disparities are accurate, pixels in the right image can move a disparity-length distance, and find their counterparts with the same intensity in the left image. The objective thus becomes minimizing a reconstruction loss between the synthesized and real left images.

For example, Ye \textit{et al.} \cite{ye2017self} proposed a Siamese model to estimate disparity and synthesize image on both left and right endoscopic images, and then trained the model by maximizing the similarity between synthesized and real stereo image pairs.
Based on Ye \textit{et al.}\cite{ye2017self}, Huang \textit{et al.}\cite{huang2021self} additionally introduced a discriminator to distinguish the synthesized images from real ones, and to play the min-max game with the disparity estimator.
However, such pixel-wise reconstruction loss works well only if the intensity value of matching pixels is consistent, which is hardly the case for stereo endoscopic images because of a strong non-Lambertian reflection caused by organs and blood.
In addition to brightness inconsistency, there may not be enough texture/features on the organ surfaces to find corresponding pixels.

\begin{figure*}[t]
	\centering
	\includegraphics[width=0.95\linewidth]{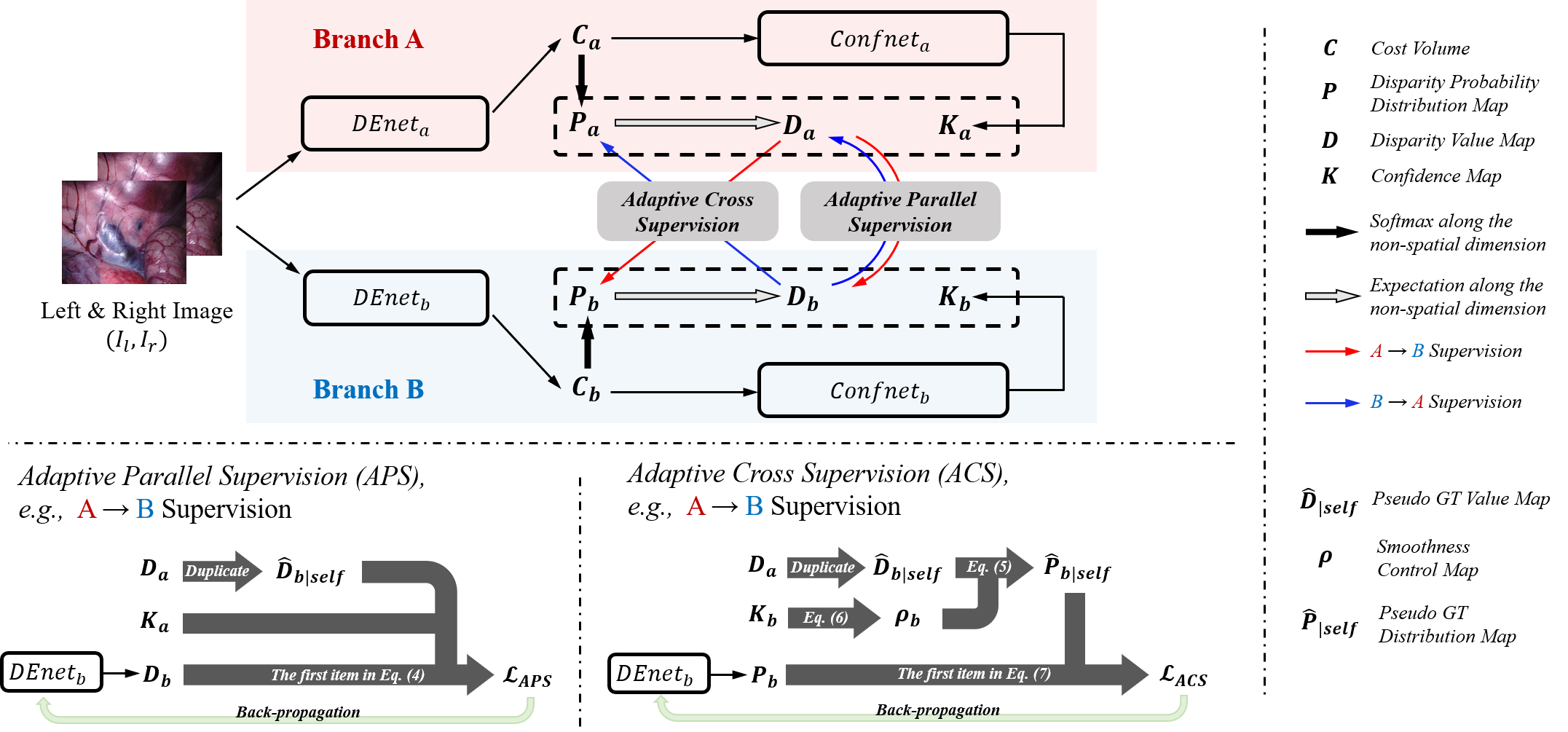}
	\caption{Illustration of dual-branch CNN's architecture and its self-supervised learning by the two adaptive bidirectional supervisions, i.e., APS and ACS.}
	\vspace{-0.4cm}
	\label{fig:two_branch}
\end{figure*}

\subsection{Semi-supervised Depth/Disparity Estimation}
Semi-supervised stereo matching combines the two above-mentioned learning fashions, and exploits labeled and unlabeled samples jointly.

Smolyanskiy \textit{et al.}\cite{smolyanskiy2018importance} extended the fully-supervised GCNet \cite{kendall2017end} by additionally optimizing a self-supervised reconstruction loss like MonoDepth \cite{godard2017unsupervised}. This method enjoyed the strengths of two learning fashions, but also shared the flaws discussed in Sec. \ref{SEC2.1} and \ref{SEC2.2}.
To prevent inheriting the limitations of both in the joint optimization, Baek~\textit{et al.}\cite{baek2022semi} 
utilized images with sparse ground truth to train two independent branches using the supervised and unsupervised losses respectively. Between the two branches, they proposed a mutual distillation-based loss to distill depth information from each other.
This approach combines the advantages of supervised learning for recovering object edges well and unsupervised learning for predicting depth in areas without ground truth supervision (such as the sky and transparent objects). However, their method still requires a large number of images with sparse labels, which are easily obtainable in natural scenes, but quite challenging for in-vivo environments.

Some methods \cite{luo2019details,tosi2019learning,tonioni2019unsupervised} proposed to generate pseudo labels of unlabeled samples by traditional stereo matching for training models.
For example, Luo \textit{et al.}\cite{luo2019details} first utilized AD-Census \cite{mei2011building} to estimate disparities of unlabeled samples, and then shortlisted the reliable predictions by assessing their confidence according to \cite{poggi2017quantitative}.
Tonioni \textit{et al.}\cite{tonioni2019unsupervised} adopted a similar idea in domain adaption. They used SGM\cite{hirschmuller2007stereo} to predict pseudo disparities of unlabeled samples of another domain, and estimate a corresponding confidence map to suppress errors in the prediction. Then the pseudo labels are reweighted to finetune the model on the target dataset.
However, the pseudo labels generated by traditional methods are typically inaccurate due to low-textured surfaces in endoscopic images, and thus become very sparse after low-confidence filtering, which limits the efficacy of unlabeled data.

%\vspace{-10pt}

In semi-supervised learning, the effectiveness of Teacher Student Network (TSN) \cite{laine2016temporal} has been demonstrated in various computer vision tasks \cite{li2018semi,cui2019semi}.
However, TSN cannot handle medical data because it heavily relies on the predictions of the teacher model, and a teacher model trained with a small amount of labeled data can easily overfit, leading to poor quality of pseudo labels and thus degrading the performance of the student model.
To solve this, Wang \textit{et al.}~\cite{wang2022rethinking} proposed a Bayesian deep learning architecture based on TSN, where both labeled and unlabeled data can be utilized to estimate the joint distribution, alleviating overfitting problem caused by only using labeled data in the early training stage.
In our previous work \cite{shi2021semi}, we made two distinctive improvements to the original TSN: we trained an enhanced teacher model to predict mostly-correct pseudo labels by using an additional adversarial learning; we learned a confidence network to lower the weights of the remaining incorrect predictions in the loss calculation. However, the teachers in both Wang \textit{et al.}~\cite{wang2022rethinking} and our previous work are trained in isolation, and receives no feedback from the students. This makes the knowledge flow unidirectional from the teacher to student, in which the TSN accuracy becomes teacher-dependent, and the learning of student model could be suboptimal due to the stagnant quality of pseudo labels.

\vspace{-0.1cm}
\section{Method}
%\vspace{-0.15cm}
%Figure \ref{fig:two_branch} illustrates our unified framework which is a dual-branch CNN, where each branch consists of a Disparity Estimation network (i.e., $ DEnet$) and a Confidence network (i.e., $ Confnet$). 
In this section, we first introduce the architecture of dual-branch CNN in Sec.~\ref{SEC3.1}, 
and then detail the self-supervised learning via APS and ACS in Sec.~\ref{SEC3.2}, 
and the fully-supervised learning of each individual branch in Sec.~\ref{SEC3.3}, 
and give implementation and training details in Sec.~\ref{SEC3.4}.

\vspace{-0.3cm}
\subsection{Architecture of Dual-branch Network}
\label{SEC3.1}
%\\ \hspace*{\fill} \\

\subsubsection{Dual-Branch Architecture} The top part of Fig. \ref{fig:two_branch} illustrates the dual-branch CNN, and each branch consists of two key networks, i.e., a Disparity Estimation network (i.e., $ DEnet $) and a Confidence network (i.e., $ Confnet $). 
For clarity, we use two subscripts $ a $ and $ b $ to distinguish the networks and predictions in different branches, and the capital letters $ A $ and $ B $ to distinguish the two branches.

Given a pair of rectified left-right images $ (I_{l}, I_{r}) $ with the size of $ H \times W $, Branch A predicts three maps of disparity probability distribution, disparity value, and confidence, which are denoted as $ P_a $, $ D_a $, and $ K_a $, respectively. Also, Branch B predicts $ P_b $, $ D_b $, and $ K_b $, simultaneously.

In the following, we detail the architecture of $ DEnet $ and $ Confnet $, and list all layer specifications in Table \ref{notion}.
We discard the subscripts for a general description since the two branches share the identical architecture (but with a different weight initialization).

\subsubsection{Disparity Estimation Network (DEnet)} 
\begin{figure}[t]
	\centering
	\includegraphics[width=\linewidth]{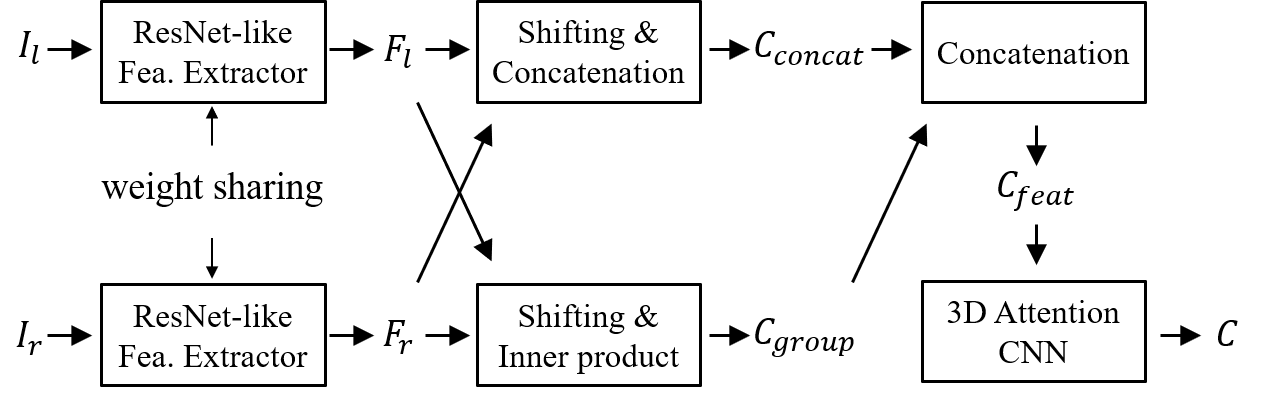}
	\caption{The illustrations of $ DEnet $'s architecture, which contains a weight-sharing ResNet-like feature extractor and a 3D Attention CNN. The 3D feature volume $ C_{feat}$ is constructed based on $ C_{concat} $ and $ C_{group} $, and then converted to a cost volume $ C $.}
	\vspace{-0.5cm}
	\label{fig:DENET_architecture}
\end{figure}

Fig.~\ref{fig:DENET_architecture} illustrates the architecture of $ DEnet $ which utilizes a weight-sharing ResNet-like network to extract features from $ I_{l} $ and $ I_{r} $, yielding two downscaled feature maps $ F_{l} $ and $ F_{r} $ with the size of $ 320\times H/4\times W/4 $.
Specifically, $ F_{l} $ is obtained by concatenating feature maps from the last three residual 2d blocks of the feature extractor in Table \ref{notion} , and so is $ F_{r} $.
After that, $ F_{l} $ and $ F_{r} $ are fused to construct a feature volume $ C_{feat} $ with the size of $ 64\times H/4\times W/4\times S/4 $ where $ S $ denotes the maximum disparity, and thus the last dimension of $ C_{feat} $ defines a disparity searching space (also downscaled) with a disparity range from $0$ to $ S-1 $ pixels.

As illustrated in Fig.~\ref{fig:DENET_architecture}, $ {C}_{feat} $ is composed of two intermediates, i.e., concatenation volume $ {C}_{concat} $ and group-wise correlation volume $ {C}_{group} $. $ {C}_{concat} $ is obtained by shifting and concatenation at different disparity levels. Specifically, the two feature maps $ F_{l} $ and $ F_{r} $ are first channel-compressed using two weight-sharing convolutional layers, which decreases their channel number to 12. At each disparity level $ s $, the channel-compressed right feature map is right-ward translated $ s $ pixels, and then concatenated with the channel-compressed left feature map. Therefore, $ {C}_{concat} $ at position $ (x,y,s) $ is calculated as follows:

\begin{equation}
	\begin{aligned}
		{C}_{concat}(x,y,s)={concat}(&{convs} (F_l(x,y)), \\
		&{convs} (F_r(x-s,y)))
		\label{equ:concat}
	\end{aligned}
\end{equation}
where $ s=0,1,...,S/4-1 $ since each spatial dimension is downscaled by four times in the feature maps, $ {concat}() $ means feature concatenation along channel, and $ {convs}() $ means two sequential convolutions for channel compression. $ {C}_{concat} $ is thus with the size of $ 24\times H/4\times W/4\times S/4 $.

To ease the difficulty of fully revealing the matching relations from the single feature volume $ {C}_{concat} $, we also compute a group-wise correlation volume $ {C}_{group} $ to provide backup feature relation information.

$ {C}_{group} $ is obtained by shifting and inner product at different disparity levels. Specifically, we evenly divide $ F_{l} $ into $ N_g $ segments along channel dimension, and the $ g $-th divided feature map $ F^g_l $ is with the size of $ 320/N_g\times H/4\times W/4 $. The same procedure is also performed on $ F^g_r $.

We encode the feature relation between the $ g $-th segments from $ F_l $ and $ F_r $ into the $ g $-th channel of $ C_{group} $, and compute the $ g $-th channel value of $ C_{group} $ at position $ (x,y,s) $ as follows:

\begin{equation}
	\begin{aligned}
		{C}_{group}(g,x,y,s)= \frac{N_{g}}{320}\langle F_l^g(x,y),F_r^g(x-s,y)\rangle
		\label{equ:gwc}
	\end{aligned}
\end{equation}
where $ g=0,1,...,N_g-1 $, and $\langle \cdot, \cdot \rangle$ represents the inner product between two feature vectors. $ N_g $ is set to 40 in our experiments, and $ C_{group} $ is thus with the size of $ 40\times H/4\times W/4\times S/4 $. Finally, we concatenate both $ C_{concat} $ and $ C_{group} $, and get the desired feature volume $ C_{feat} $ with the size of $ 64\times H/4\times W/4\times S/4 $, that is, $ C_{feat}=concat(C_{concat}, C_{group}) $.

\begingroup
\renewcommand{\arraystretch}{1.2}
\begin{table}[]
	\caption{Structure details of $ DEnet $ and $ Confnet $. H and W represent the height and the width of the input image, respectively. k is the kernel size in the operator, n is the repeating times of the operator, s is the downscaling size of the operator. If not specified, a batch normalization and ReLU are adopted. The mark * denotes no ReLU, and the mark ** denotes convolution only. }
	%	\small
	%	\footnotesize 
	\scriptsize
	\begin{tabular*}{0.99\linewidth}{c|c|c|c|c|c}
		%		\toprule[1.5pt]
		\thickhline
		
		\multicolumn{1}{l|}{} &
		\multicolumn{1}{l|}{\multirow{1}{*}{Input}} & \multicolumn{1}{l|}{\multirow{1}{*}{Operator}} & \multicolumn{1}{c|}{\multirow{1}{*}{k}}  &  \multicolumn{1}{c|}{\multirow{1}{*}{n}}  & \multicolumn{1}{c}{\multirow{1}{*}{s}}                                           \\ \hline

		%		\multicolumn{7}{c}{$ DENet $} \\ \hline

		\multicolumn{1}{l|}{\multirow{11}{*}{\rotatebox{90}{$ DEnet $}}} &
		\multicolumn{1}{l|}{\multirow{1}{*}{3$\times$H$\times$W ($I_l$/$I_r$)}} & \multicolumn{1}{l|}{\multirow{1}{*}{Feature Extractor}} & \multicolumn{1}{c|}{\multirow{1}{*}{}}   & \multicolumn{1}{c|}{\multirow{1}{*}{}}   & \multicolumn{1}{c}{\multirow{1}{*}{}}                                                         \\
		\cline{2-6}
		
		\multicolumn{1}{l|}{} &
		\multicolumn{1}{l|}{\multirow{2}{*}{320$\times$H/4$\times$W/4 ($F_l$/$F_r$)}} & \multicolumn{1}{l|}{\multirow{1}{*}{shift \& concat.}} & \multicolumn{1}{c|}{\multirow{1}{*}{}}   & \multicolumn{1}{c|}{\multirow{1}{*}{}}   & \multicolumn{1}{c}{\multirow{1}{*}{}}                                                         \\
		\cline{3-6}

		\multicolumn{1}{l|}{} &
		\multicolumn{1}{l|}{\multirow{2}{*}{}} & 
		\multicolumn{1}{l|}{\multirow{1}{*}{shift \& inner product}} & \multicolumn{1}{c|}{\multirow{1}{*}{}}   & \multicolumn{1}{c|}{\multirow{1}{*}{}}   & \multicolumn{1}{c}{\multirow{1}{*}{}}                                                         \\
		\cline{2-6}
		
		\multicolumn{1}{l|}{} &
		\multicolumn{1}{l|}{\multirow{1}{*}{24$\times$H/4$\times$W/4$\times$S/4 \tiny$ \rm(C_{concat}) $}} & 
		\multicolumn{1}{l|}{\multirow{2}{*}{concat.}} & \multicolumn{1}{c|}{\multirow{1}{*}{}}   & \multicolumn{1}{c|}{\multirow{1}{*}{}}   & \multicolumn{1}{c}{\multirow{1}{*}{}}                                                         \\
		\cline{2-2}

		\multicolumn{1}{l|}{} &
		\multicolumn{1}{l|}{\multirow{1}{*}{40$\times$H/4$\times$W/4$\times$S/4 \tiny$ \rm(C_{group}) $}} & 
		\multicolumn{1}{l|}{\multirow{2}{*}{}} &  \multicolumn{1}{c|}{\multirow{1}{*}{}}   & \multicolumn{1}{c|}{\multirow{1}{*}{}}   & \multicolumn{1}{c}{\multirow{1}{*}{}}                                                         \\
		\cline{2-6}	
		
		\multicolumn{1}{l|}{} &
		\multicolumn{1}{l|}{\multirow{1}{*}{64$\times$H/4$\times$W/4$\times$S/4 \tiny$ \rm(C_{feat}) $}} & 
		\multicolumn{1}{l|}{\multirow{1}{*}{conv3d}} & \multicolumn{1}{c|}{\multirow{1}{*}{3}}   &  \multicolumn{1}{c|}{\multirow{1}{*}{2}}   & \multicolumn{1}{c}{\multirow{1}{*}{1}}                                                         \\
		\cline{2-6}	
		
		\multicolumn{1}{l|}{} &
		\multicolumn{1}{l|}{\multirow{1}{*}{32$\times$H/4$\times$W/4$\times$S/4}} & 
		\multicolumn{1}{l|}{\multirow{1}{*}{residual 3d block}} & \multicolumn{1}{c|}{\multirow{1}{*}{3}}   & \multicolumn{1}{c|}{\multirow{1}{*}{1}}   & \multicolumn{1}{c}{\multirow{1}{*}{1}}                                                         \\
		\cline{2-6}	
		
		\multicolumn{1}{l|}{} &
		\multicolumn{1}{l|}{\multirow{1}{*}{32$\times$H/4$\times$W/4$\times$S/4}} & 
		\multicolumn{1}{l|}{\multirow{1}{*}{3D Attention U-Net}} & \multicolumn{1}{c|}{\multirow{1}{*}{}}    & \multicolumn{1}{c|}{\multirow{1}{*}{3}}   & \multicolumn{1}{c}{\multirow{1}{*}{}}                                                         \\
		\cline{2-6}	
		
		\multicolumn{1}{l|}{} &
		\multicolumn{1}{l|}{\multirow{1}{*}{32$\times$H/4$\times$W/4$\times$S/4}} & 
		\multicolumn{1}{l|}{\multirow{1}{*}{conv3d**}} & \multicolumn{1}{c|}{\multirow{1}{*}{3}}   & \multicolumn{1}{c|}{\multirow{1}{*}{1}}   & \multicolumn{1}{c}{\multirow{1}{*}{1}}                                                         \\
		\cline{2-6}	
		
		\multicolumn{1}{l|}{} &
		\multicolumn{1}{l|}{\multirow{1}{*}{1$\times$H/4$\times$W/4$\times$S/4}} & 
		\multicolumn{1}{l|}{\multirow{1}{*}{upsample \& squeeze}} & \multicolumn{1}{c|}{\multirow{1}{*}{}}   & \multicolumn{1}{c|}{\multirow{1}{*}{}}   & \multicolumn{1}{c}{\multirow{1}{*}{}}                                                         \\
		\cline{2-6}	
		
		\multicolumn{1}{l|}{} &
		\multicolumn{1}{l|}{\multirow{1}{*}{H$\times$W$\times$S ($ C $)}} & 
		\multicolumn{1}{l|}{\multirow{1}{*}{Output}} & \multicolumn{1}{c|}{\multirow{1}{*}{}}   & \multicolumn{1}{c|}{\multirow{1}{*}{}}   & \multicolumn{1}{c}{\multirow{1}{*}{}}                                                         \\
		\hline

		\multicolumn{1}{l|}{\multirow{3}{*}{\rotatebox{90}{\tiny$Confnet$}}} &
		\multicolumn{1}{l|}{\multirow{1}{*}{H$\times$W$\times$S ($ C $)}} & \multicolumn{1}{l|}{\multirow{1}{*}{conv2d}} & \multicolumn{1}{c|}{\multirow{1}{*}{3}}   & \multicolumn{1}{c|}{\multirow{1}{*}{1}}   & \multicolumn{1}{c}{\multirow{1}{*}{1}}                                                         \\
		\cline{2-6}	
		
		\multicolumn{1}{l|}{} &
		\multicolumn{1}{l|}{\multirow{1}{*}{H$\times$W$\times$S/3}} & \multicolumn{1}{l|}{\multirow{1}{*}{conv2d**}} & \multicolumn{1}{c|}{\multirow{1}{*}{1}}   & \multicolumn{1}{c|}{\multirow{1}{*}{1}}   & \multicolumn{1}{c}{\multirow{1}{*}{1}}                                                         \\
		\cline{2-6}	
		
		\multicolumn{1}{l|}{} &
		\multicolumn{1}{l|}{\multirow{1}{*}{H$\times$W$\times$1 ($ K $)}} & \multicolumn{1}{l|}{\multirow{1}{*}{Output}} & \multicolumn{1}{c|}{\multirow{1}{*}{}}   & \multicolumn{1}{c|}{\multirow{1}{*}{}}   & \multicolumn{1}{c}{\multirow{1}{*}{}}         \\
		\hline

		\multicolumn{1}{l|}{\multirow{8}{*}{\rotatebox{90}{ResNet-like Fea. Extractor}}} &
		
		\multicolumn{1}{l|}{\multirow{1}{*}{3$\times$H$\times$W}} & \multicolumn{1}{l|}{\multirow{1}{*}{conv2d}} & \multicolumn{1}{c|}{\multirow{1}{*}{3}}   &  \multicolumn{1}{c|}{\multirow{1}{*}{1}}   & \multicolumn{1}{c}{\multirow{1}{*}{2}}                                                         \\
		\cline{2-6}
		
		%		\multicolumn{1}{l|}{}                            & \multicolumn{1}{l}{}                                                                            \\ 
		
		\multicolumn{1}{l|}{} &
		\multicolumn{1}{l|}{\multirow{1}{*}{32$\times$H/2$\times$W/2}} & \multicolumn{1}{l|}{\multirow{1}{*}{conv2d}} & \multicolumn{1}{c|}{\multirow{1}{*}{3}}   & \multicolumn{1}{c|}{\multirow{1}{*}{2}}   & \multicolumn{1}{c}{\multirow{1}{*}{1}}                                                         \\
		\cline{2-6}
		\multicolumn{1}{l|}{} &
		\multicolumn{1}{l|}{\multirow{1}{*}{32$\times$H/2$\times$W/2}} & \multicolumn{1}{l|}{\multirow{1}{*}{residual 2d block}} & \multicolumn{1}{c|}{\multirow{1}{*}{3}}   & \multicolumn{1}{c|}{\multirow{1}{*}{3}}   & \multicolumn{1}{c}{\multirow{1}{*}{1}}                                                         \\
		\cline{2-6}
		
		\multicolumn{1}{l|}{} &
		\multicolumn{1}{l|}{\multirow{1}{*}{32$\times$H/2$\times$W/2}} & \multicolumn{1}{l|}{\multirow{1}{*}{residual 2d block}} & \multicolumn{1}{c|}{\multirow{1}{*}{3}}   & \multicolumn{1}{c|}{\multirow{1}{*}{16}}   & \multicolumn{1}{c}{\multirow{1}{*}{2}}                                                         \\
		\cline{2-6}
		
		\multicolumn{1}{l|}{} &
		\multicolumn{1}{l|}{\multirow{1}{*}{64$\times$H/4$\times$W/4 (a)}} & \multicolumn{1}{l|}{\multirow{1}{*}{residual 2d block}} & \multicolumn{1}{c|}{\multirow{1}{*}{3}}   & \multicolumn{1}{c|}{\multirow{1}{*}{3}}   & \multicolumn{1}{c}{\multirow{1}{*}{1}}                                                         \\
		\cline{2-6}
		
		\multicolumn{1}{l|}{} &
		\multicolumn{1}{l|}{\multirow{1}{*}{128$\times$H/4$\times$W/4 (b)}} & \multicolumn{1}{l|}{\multirow{1}{*}{residual 2d block}} & \multicolumn{1}{c|}{\multirow{1}{*}{3}}  & \multicolumn{1}{c|}{\multirow{1}{*}{3}}   & \multicolumn{1}{c}{\multirow{1}{*}{1}}                                                         \\
		\cline{2-6}
		
		\multicolumn{1}{l|}{} &
		\multicolumn{1}{l|}{\multirow{1}{*}{128$\times$H/4$\times$W/4}} & \multicolumn{1}{l|}{\multirow{1}{*}{concat. with (a) \& (b)
		}} & \multicolumn{1}{l|}{\multirow{1}{*}{}}   &  \multicolumn{1}{l|}{\multirow{1}{*}{}}   & \multicolumn{1}{l}{\multirow{1}{*}{}}                                                         \\
		
		\cline{2-6}
		
		\multicolumn{1}{l|}{\multirow{2}{*}{}} &
		\multicolumn{1}{l|}{\multirow{1}{*}{320$\times$H/4$\times$W/4}} & \multicolumn{1}{l|}{\multirow{1}{*}{Output}} & 
		\multicolumn{1}{l|}{\multirow{2}{*}{}}   & \multicolumn{1}{l|}{\multirow{2}{*}{}}   & \multicolumn{1}{l}{\multirow{2}{*}{}}                                                         \\
		
		\hline

		\multicolumn{1}{l|}{\multirow{11}{*}{\rotatebox{90}{3D Attention U-Net}}} &

		\multicolumn{1}{l|}{\multirow{1}{*}{32$\times$H/4$\times$W/4$\times$S/4 (x)}} & \multicolumn{1}{l|}{\multirow{1}{*}{conv3d}} & \multicolumn{1}{c|}{\multirow{1}{*}{3}}   &  \multicolumn{1}{c|}{\multirow{1}{*}{1}}   & \multicolumn{1}{c}{\multirow{1}{*}{2}}                                                         \\
		\cline{2-6}
		
		\multicolumn{1}{l|}{} &
		\multicolumn{1}{l|}{\multirow{1}{*}{64$\times$H/8$\times$W/8$\times$S/8}} & \multicolumn{1}{l|}{\multirow{1}{*}{conv3d}} & \multicolumn{1}{c|}{\multirow{1}{*}{3}}   &  \multicolumn{1}{c|}{\multirow{1}{*}{1}}   & \multicolumn{1}{c}{\multirow{1}{*}{1}}                                                         \\
		\cline{2-6}
		
		\multicolumn{1}{l|}{} &
		\multicolumn{1}{l|}{\multirow{1}{*}{64$\times$H/8$\times$W/8$\times$S/8 (y)}} & \multicolumn{1}{l|}{\multirow{1}{*}{conv3d}} & \multicolumn{1}{c|}{\multirow{1}{*}{3}}   &  \multicolumn{1}{c|}{\multirow{1}{*}{1}}   & \multicolumn{1}{c}{\multirow{1}{*}{2}}                                                         \\
		\cline{2-6}
		
		\multicolumn{1}{l|}{} &
		\multicolumn{1}{l|}{\multirow{1}{*}{128$\times$H/16$\times$W/16$\times$S/16}} & \multicolumn{1}{l|}{\multirow{1}{*}{conv3d}} & \multicolumn{1}{c|}{\multirow{1}{*}{3}}   &  \multicolumn{1}{c|}{\multirow{1}{*}{1}}   & \multicolumn{1}{c}{\multirow{1}{*}{1}}                                                         \\
		\cline{2-6}
		
		\multicolumn{1}{l|}{} &
		\multicolumn{1}{l|}{\multirow{1}{*}{128$\times$H/16$\times$W/16$\times$S/16}} & \multicolumn{1}{l|}{\multirow{1}{*}{channel attention}} & \multicolumn{1}{c|}{\multirow{1}{*}{}}   &  \multicolumn{1}{c|}{\multirow{1}{*}{}}   & \multicolumn{1}{c}{\multirow{1}{*}{}}                                                         \\
		\cline{2-6}
		
		\multicolumn{1}{l|}{} &
		\multicolumn{1}{l|}{\multirow{1}{*}{128$\times$H/16$\times$W/16$\times$S/16}} & \multicolumn{1}{l|}{\multirow{1}{*}{deconv3d*}} & \multicolumn{1}{c|}{\multirow{1}{*}{3}}   & \multicolumn{1}{c|}{\multirow{1}{*}{1}}   & \multicolumn{1}{c}{\multirow{1}{*}{2}}                                                         \\
		\cline{2-6}
		
		\multicolumn{1}{l|}{} &
		\multicolumn{1}{l|}{\multirow{1}{*}{64$\times$H/8$\times$W/8$\times$S/8}} & \multicolumn{1}{l|}{\multirow{1}{*}{add with (y) then Relu}} & \multicolumn{1}{c|}{\multirow{1}{*}{}}   & \multicolumn{1}{c|}{\multirow{1}{*}{}}   & \multicolumn{1}{c}{\multirow{1}{*}{}}                                                         \\ 
		
		\cline{2-6}
		
		\multicolumn{1}{l|}{} &
		\multicolumn{1}{l|}{\multirow{1}{*}{64$\times$H/8$\times$W/8$\times$S/8}} & \multicolumn{1}{l|}{\multirow{1}{*}{deconv3d*}} & \multicolumn{1}{c|}{\multirow{1}{*}{3}}   &  \multicolumn{1}{c|}{\multirow{1}{*}{1}}   & \multicolumn{1}{c}{\multirow{1}{*}{2}}                                                         \\
		\cline{2-6}
		
		\multicolumn{1}{l|}{} &
		\multicolumn{1}{l|}{\multirow{1}{*}{32$\times$H/4$\times$W/4$\times$S/4}} & \multicolumn{1}{l|}{\multirow{1}{*}{add with (x) then Relu}} & \multicolumn{1}{c|}{\multirow{1}{*}{}}   & \multicolumn{1}{c|}{\multirow{1}{*}{}}   & \multicolumn{1}{c}{\multirow{1}{*}{}}                                                         \\
		
		\cline{2-6}
		
		\multicolumn{1}{l|}{} &
		\multicolumn{1}{l|}{\multirow{1}{*}{32$\times$H/4$\times$W/4$\times$S/4}} & \multicolumn{1}{l|}{\multirow{1}{*}{conv3d}} & \multicolumn{1}{c|}{\multirow{1}{*}{3}}   & \multicolumn{1}{c|}{\multirow{1}{*}{1}}   & \multicolumn{1}{c}{\multirow{1}{*}{1}}                                                         \\
		\cline{2-6}
		
		\multicolumn{1}{l|}{} &
		\multicolumn{1}{l|}{\multirow{1}{*}{32$\times$H/4$\times$W/4$\times$S/4}} & \multicolumn{1}{l|}{\multirow{1}{*}{Output}} & \multicolumn{1}{c|}{\multirow{1}{*}{}}   & \multicolumn{1}{c|}{\multirow{1}{*}{}}   & \multicolumn{1}{c}{\multirow{1}{*}{}}                                                         \\
		
		\thickhline
		%	\bottomrule[1.5pt]
	\end{tabular*}
	
	\label{notion}
\end{table}
\endgroup

Next, a 3D Attention CNN is used to aggregate and regularize $ C_{feat} $ into a cost volume $ C $ (CV) with the size of $ H\times W\times S$.
The 3D Attention CNN consists of three cascaded U-Nets~\cite{ronneberger2015u} (refer to the 3D Attention U-Net in Table~\ref{notion}). 
In each U-Net, the channel attention mechanism is embedded after the last encoding layer to enhance inter-dependency of features at different disparity levels.

After that, a softmax function \cite{kendall2017end} converts $ C $ to a map of disparity probability distribution $ {P} $, which is calculated as follows:

\begin{equation}
	\small
	P(x,y,s) = \frac{\exp(-C(x,y,s))}{\sum_{s=0}^{S-1}\exp(-C(x,y,s))}
	\label{equ:softmax}
\end{equation}
%\end{small}
where $ s=0,1,...,S-1 $. $ P(x,y,s) $ indicates a probability of the disparity being $ s $ pixels at the position $ (x,y) $. 

Finally, a map of disparity value $ {D} $ is calculated as the expectation of $ P $, which is formulated in Eq. (\ref{equ:disp_regression}).

\begin{small}
\begin{equation}
	{D(x,y)} = \sum_{s=0}^{S-1} s \times P(x,y,s)
	\label{equ:disp_regression}
\end{equation}
\end{small}

\vspace{-0.1cm}
\subsubsection{Confidence Network (Confnet)} $Confnet$ follows $DEnet$, and estimates how accurate the prediction of $ DEnet $ is. 
As shown in Table \ref{notion}, $ Confnet $ takes the predicted CV as an input, and utilizes two convolutional layers and a batch normalization to generate a confidence map $ K $, whose values are normalized via a sigmoid into a range from 0 to 1. Lower values in $ K $ indicate higher possibilities that errors occur on the corresponding spatial positions in $ D $.
%` which is converted from $ C $.

\subsection{Self-supervised Learning via APS and ACS}
\label{SEC3.2}

%After a forward processing on an unlabeled sample, the two self-supervisions (i.e., APS and ACS) make the two branches teach each other mutually.
%\\ \hspace*{\fill} \\

%\noindent
\subsubsection{Adaptive Parallel Supervision (APS)} As shown in the bottom-left part of Fig. \ref{fig:two_branch} where the knowledge flows from \emph{Branch A to Branch B}, APS treats $ D_a $ from Branch A as a pseudo ground-truth (GT) value map, denoted as $ \hat{D}_{b|self} $, to guide the learning of Branch B by minimizing the smooth L1 loss (see Eq.~(\ref{equ:smooth L1})) between $ \hat{D}_{b|self} $ and $ D_{b} $.
% The smooth L1 loss is computed as follows:

\begin{small}
\begin{equation}
	{smooth}_{L_{1}}(x)=\left\{\begin{array}{ll}0.5 x^{2}, & \text { if }|x|<1 \\|x|-0.5, & \text { otherwise }\end{array}\right.
	\label{equ:smooth L1}
\end{equation}
\end{small}

To increase the reliability of $ \hat{D}_{b|self} $, $ K_{a}$ is utilized in a re-weighting strategy to suppress the possible wrong predictions in $ \hat{D}_{b|self} $.
% $ M_a $ is calculated as follows:

%\begin{small}
%\begin{equation}
%	\begin{aligned}
%		M_{a}(x,y)=\left\{\begin{array}{ll}1, & \text { if }K_{a}(x,y) \ge \Omega \\0, & \text { otherwise }\end{array}\right.
%	\end{aligned}
%	\label{equ:mask}
%\end{equation}
%\end{small}
%where $ \Omega $ is a confidence threshold and set to 0.8 in our experiments. 
%%The resulting $ {M_{a}} \in [0,1] $ can be used to suppress the unreliable pseudo labels provided to Branch B.

On the opposite direction from \emph{Branch B to Branch A}, APS also minimizes the smooth L1 loss by treating $ D_b $ as a pseudo GT value map, denoted as $ \hat{D}_{a|self} $, to guide the learning of Branch A, and meanwhile utilizes $ K_b $ to suppress the unreliable predictions in $ \hat{D}_{a|self} $.

Thus, the final bidirectional APS loss $ \mathcal{L}_{APS} $ considers both parallel directions of knowledge flow between the two branches, and is formulated as follows:

%\begin{footnotesize}
\begin{equation}
	\footnotesize\begin{aligned}
		\mathcal{ L}_{APS} = \frac{1}{HW} \sum_{x,y}& \Bigl\{ \underbrace{K_a(x,y) \cdot smooth_{L_1}(D_b(x,y)-\hat{D}_{b|self}(x,y)) } & \\
		& ~~~~~~~~~~~~~~~~~~~~~~~~~~~A \rightarrow B & \\ 
		+ & \underbrace{ K_b(x,y) \cdot smooth_{L_1}(D_a(x,y)-\hat{D}_{a|self}(x,y)) } \Bigr\} \\ 
		& ~~~~~~~~~~~~~~~~~~~~~~~~~~B \rightarrow A 
	\end{aligned}
	\label{equ:APS_loss}
\end{equation}
%\end{footnotesize}
where $ \hat{D}_{b|self}=D_a $, $ \hat{D}_{a|self}=D_b $.

It is worth noting that the gradients derived from minimizing the first and second item in Eq. (\ref{equ:APS_loss}) only update the weights of $ DEnet_b $ and $ DEnet_a $, respectively.
%\\ \hspace*{\fill} \\
%\noindent

\vspace{0.2cm}
\subsubsection{Adaptive Cross Supervision (ACS)} 
%As we described in Sec.~\ref{SEC3.1}, it is direct objective of each branch to model the distribution $ P $, and the disparity value $ D $ is just its expectation. 
If we only constrain the disparity value, there could be an abnormal solution where an erroneous distribution happens to derive a correct disparity value, therefore, the network learns questionable feature relations at some disparity levels. 
To avoid this, we additionally constrain the predicted distribution map in ACS.

As shown in the bottom-right part of Fig. \ref{fig:two_branch} where the knowledge flows from \emph{Branch A to Branch B},
ACS constructs a pseudo GT distribution map $ \hat{P}_{b|self} $ from $ \hat{D}_{b|self} $ to supervise the learning of Branch B. To this end, we create three rules to the construction procedure of GT distribution:

(1) The constructed GT disparity probability distribution obeys a unimodal distribution;

(2) The peak of the unimodal distribution locates around the corresponding true disparity value;

(3) The peak distribution is wider if the disparity value is more difficult to estimate (lower prediction confidence).

The first two rules are motivated by a fact that the disparity probability distribution actually reflects how well a pixel pair matches each other at different disparity levels. 
Therefore, the matching degree should reach its maximum around the true disparity, and decreases as the disparity level gets away from the true value. The third rule is inspired by the label-softening technique \cite{muller2019does} which is widely employed in the classification task. A hard classification label is better to be softened from a one-hot vector to a distribution if this particular class is easily confused with others. In our case, we smooth the GT unimodal distribution of the regions with low confidence where the true disparity could be easily confused with its nearby values.

\begin{figure}[t]
	\centering
	\includegraphics[width=\linewidth]{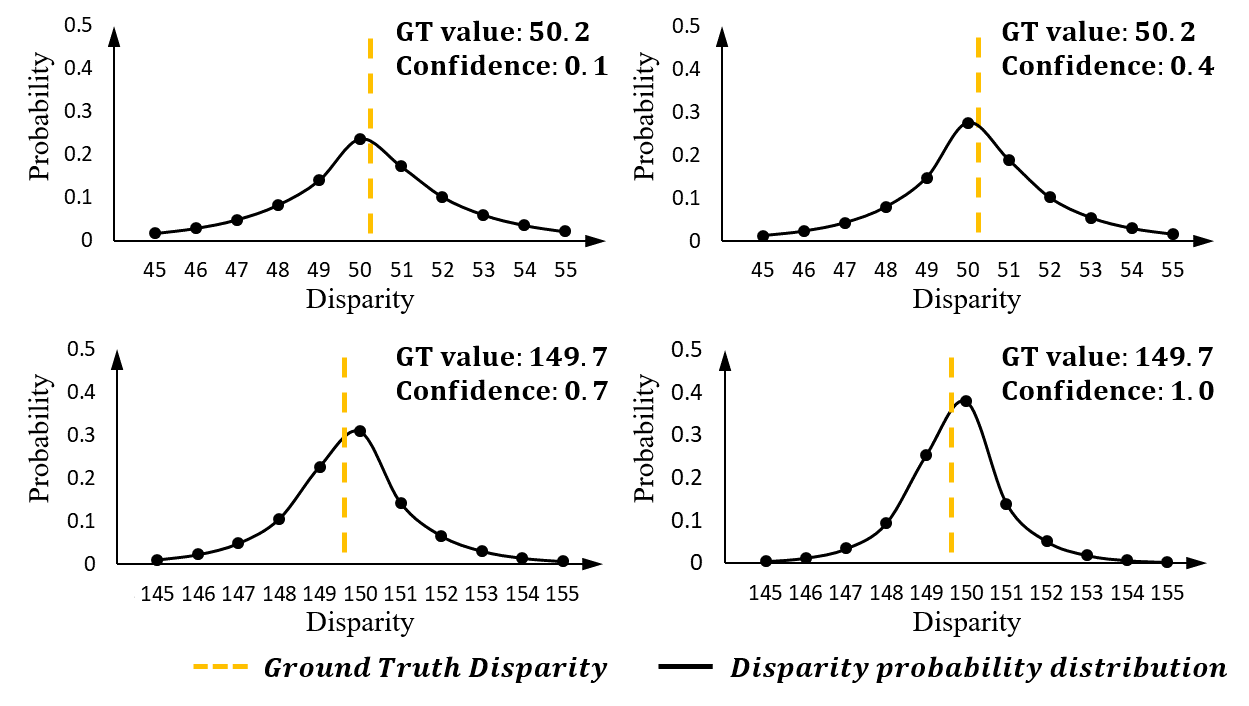}
	\caption{Four examples of unimodal disparity probability distribution generated by our designed non-parametric operation of unimodal generation. After the operation, the disparity values can be softened into distributions, where the peak aligns the GT value, and is wider and lower if the confidence decreases.}
	\vspace{-0.4cm}
	\label{fig:disp_distribution}
\end{figure}

To obey the above three rules, we design a non-parametric operation of unimodal generation (denoted as $ UG $), which inputs the GT value and confidence, and outputs a corresponding GT distribution. 
Therefore, we have $\hat{P}_{b|self}= UG(\hat{D}_{b|self}, K_b)$, and $ \hat{P}_{b|self}(x,y,s) $ is calculated as follows:

%\begin{footnotesize}
\begin{equation}
	\footnotesize
	\hat{P}_{b|self}(x,y,s)=\frac{\exp(-\left | s-\hat{D}_{b|self}(x,y) \right | \cdot \rho_b(x,y) )}{ {\textstyle \sum_{s=0}^{S-1}\exp(-\left | s-\hat{D}_{b|self}(x,y) \right | \cdot \rho_b(x,y))} }
	\label{equ:P-confidence}
\end{equation}
%\end{footnotesize}
where $ \hat{D}_{b|self}=D_a $.

According to Eq.~(\ref{equ:P-confidence}), $ \hat{P}_{b|self}(x,y,s) $ reaches its maximum if $ s $ is closest to $ \hat{D}_{b|self}(x,y) $, and decreases otherwise, which obeys the first and second rules. $ \rho_b(x,y) $ controls the smoothness of $ \hat{P}_{b|self} $, and is calculated as follows:

%\vspace{-0.1cm}
\begin{small}
\begin{equation}
	\rho_b(x,y) = \frac{1}{2-K_b(x,y)}
	\label{equ:rho}
\end{equation}
\end{small}
\vspace{-0.1cm}

According to Eq.~(\ref{equ:rho}), $ \rho_b(x,y) $ ranges from 0.5 to 1.0, and is positively correlated to the prediction confidence $ K_b(x,y) $. 
If $ K_b(x,y) $ decreases (confidence gets lower), $ \rho_b(x,y) $ decreases accordingly, and narrows the gap of the power values of Euler's number $e$ in Eq.~(\ref{equ:P-confidence}).
Consequently, the resulting distribution becomes wider, which obeys the third rule. Fig. \ref{fig:disp_distribution} presents four disparity probability distributions generated from different GT and confidence values. 

On the opposite direction from \emph{Branch B to Branch A}, ACS also constrains the probability distribution by constructing $ \hat{P}_{a|self} = UG(\hat{D}_{a|self}, K_a)$ to guide the learning of Branch A, where $ \hat{D}_{a|self} = D_b $.

Thus, the final bidirectional ACS loss $ \mathcal{L}_{ACS} $ considers both cross directions of knowledge flow between the two branches, and is formulated as follows:

\vspace{-0.2cm}
\begin{footnotesize}
\begin{equation}
	\small\begin{aligned} 
		\mathcal{L}_{ACS} = -\frac{1}{HW} \sum_{x,y} \sum^{S-1}_{s=0} \Bigl\{ & \underbrace { \hat{P}_{b|self}(x,y,s) \cdot \log P_b(x,y,s) } & \\
		& ~~~~~~~~~~~~~~~A \rightarrow B & \\ 
		+ & \underbrace{ \hat{P}_{a|self}(x,y,s) \cdot \log P_a(x,y,s)} \Bigr\} \\ 
		& ~~~~~~~~~~~~~~~B \rightarrow A 
	\end{aligned}
	\label{equ:ACS_loss}
\end{equation}
\end{footnotesize}

Similar to APS, the first and second items in Eq.~(\ref{equ:ACS_loss}) are minimized to update $ DEnet_b $ and $ DEnet_a $, respectively.

The total self-supervised loss to optimize the dual-branch network can be formulated as follows:

\begin{small}
\begin{equation}
	\small\begin{aligned}
		\mathcal{L}_{self} = \mathcal{L}_{APS} + \mathcal{L}_{ACS}
		\label{equ:dual}
	\end{aligned}
\end{equation}
\end{small}
%where the coefficient $ \lambda_{APS} $ is set to 0.1 in our experiments.

%\vspace{-0.3cm}
\subsection{Fully-Supervised Learning of Each Branch}
\label{SEC3.3}

\begin{figure}[t]
	\centering
	\includegraphics[width=0.9\linewidth]{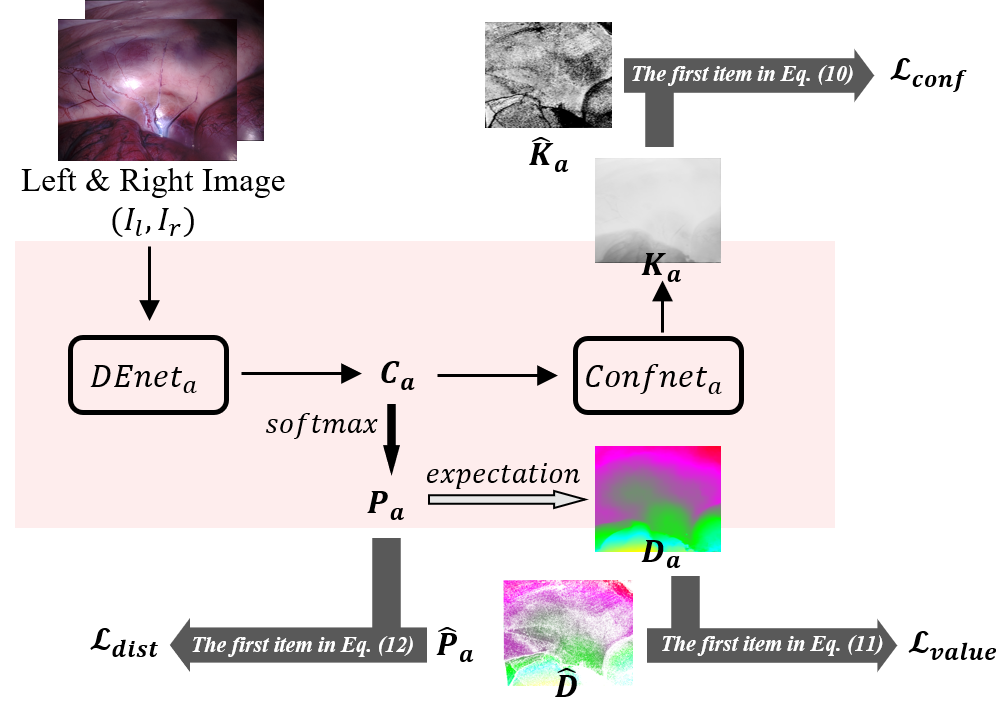}
	\caption{Illustration of the fully-supervised learning of each branch (Branch A as an example). $ \mathcal{L}_{value} $ is minimized for updating $DEnet$ only, $ \mathcal{L}_{conf} $ is minimized for updating $Confnet$ only, and $ \mathcal{L}_{dist} $ is minimized for updating $DEnet$ and $Confnet$ jointly.} 
	\vspace{-0.3cm}
	%	The probability distribution GT $ \hat{P} $ and Confidence GT $ \hat{K} $ are generated by Eq. (\ref{equ:P_hat}) and (\ref{equ:conf_generate}) respectively.}
	%		The architecture of $ Confnet $ are provided.}
	\label{fig:onebranch_super}
\end{figure}

As an example in Branch A (the same in Branch B), Fig.~\ref{fig:onebranch_super} details the fully-supervised learning of each branch on a labeled sample with its GT disparity map $ \hat{D} $ provided. We employ three losses for containing the predicted confidence $ K_a(K_b) $, disparity value $ D_a(D_b) $ and disparity probability distribution $ P_a(P_b) $. Note that, our previous work \cite{shi2021semi} only optimized the losses for confidence and disparity value, which actually trained $ DEnet $ and $ Confnet $ separately. 
In this work, we introduce an auxiliary loss for disparity probability distribution, optimizing which forces $ DEnet $ and $ Confnet $ to be trained jointly.
%\\ \hspace*{\fill} \\

\subsubsection{Confidence Constraint} 
We treat the confidence estimation as a binary classification task, and thus define a positive class of high confidence as the prediction whose error is within 3 pixels, or a negative class of low confidence otherwise. Therefore, the GT confidence map $ \hat{K_a} $ for Branch A can be generated as follows:

\begin{small}
	\begin{equation}
		\small\hat{K}_a(x,y)=\left\{\begin{array}{ll}1, & \text { if }\left|D_a(x,y)-\hat{D}(x,y)\right|<3 \\0, & \text { otherwise }\end{array}\right.
		\label{equ:conf_generate}
	\end{equation}
\end{small}

Likewise, we can also generate the GT confidence map $ \hat{K}_b $ for Branch B. The loss $ \mathcal{L}_{conf} $ based on Binary Cross Entropy loss is used to train the two confidence networks in both branches:

\vspace{-0.2cm}
\begin{footnotesize}
\begin{equation}
	\begin{aligned}
		\mathcal{L}_{conf} &= -\frac{1}{HW} \sum_{x,y} \\
		&\Bigl\{ \hat{K}_a(x,y) \cdot \log K_a(x,y) + (1-\hat{K}_a(x,y)) \cdot \log (1 - K_a(x,y))\\
		& + \hat{K}_b(x,y) \cdot \log K_b(x,y) + (1-\hat{K}_b(x,y)) \cdot \log (1 - K_b(x,y)) \Bigr\}
	\end{aligned}
\end{equation}
\end{footnotesize}
%where $ N=HW $ represents the total number of image pixels.
%\\ \hspace*{\fill} \\

\subsubsection{Disparity Value Constraint} We calculate the smooth L1 loss (see Eq.~(\ref{equ:smooth L1})) between the predicted and GT disparity value map $ \hat{D} $ to train the two disparity estimation networks in both branches. 
In addition, we use a weight related to the GT disparity value to address the potential long-tail problem. Therefore, the loss $ \mathcal{L}_{value} $ based on the disparity-aware smooth L1 loss is formulated in Eq. (\ref{equ:sup}).

%\begin{footnotesize}
\begin{equation}
	\footnotesize\begin{aligned}
		\mathcal{L}_{value} = \frac{1}{HW} \sum_{x,y} \alpha(x,y) &\cdot \bigl\{ smooth_{L_1}(D_a(x,y)-\hat{D}(x,y)) \\
		& + smooth_{L_1}(D_b(x,y)-\hat{D}(x,y)) \bigr\}
		\label{equ:sup}
	\end{aligned}
\end{equation}
%\end{footnotesize}
where $ \alpha(x,y) = {\hat{D}(x,y)} / {\max(\hat{D})} $ is defined as the normalized disparity.

\subsubsection{Disparity Probability Distribution Constraint} Similar to ACS described in Sec.~\ref{SEC3.2}, we also constrain the predicted probability distribution on each labeled sample to avoid the erroneous predictions. To this end, we utilize the unimodal generator to convert $ \hat{D} $ into two GT maps of disparity probability distribution for the two branches, that is, $ \hat{P}_a=UG(\hat{D},K_a), \hat{P}_b = UG(\hat{D}, K_b) $.

Therefore, the loss $ \mathcal{L}_{dist} $ for constraining the predicted distributions is calculated as follows:

\vspace{-0.1cm}
\begin{small}
\begin{equation}
	\begin{aligned} 
		\mathcal{L}_{dist} = -\frac{1}{HW} \sum_{x,y} \sum^{S-1}_{s=0} \Bigl\{ & \hat{P}_{a}(x,y,s) \cdot \log P_a(x,y,s) & \\
		+ & \hat{P}_{b}(x,y,s) \cdot \log P_b(x,y,s) \Bigr\}
	\end{aligned}
	\label{equ:L_dist}
\end{equation}
\end{small}

The fully-supervised loss $ \mathcal{L}_{full} $ is finally formulated in Eq.~(\ref{equ:L-full}).

\vspace{-0.1cm}
%\begin{small}
\begin{equation}
	\small\mathcal{L}_{full}=\lambda_{conf} \mathcal{L}_{conf} +\mathcal{L}_{value}+ \mathcal{L}_{dist}
	\label{equ:L-full}
\end{equation}
%\end{small}
where $ \lambda_{conf} $ denotes a weight coefficient, and is tuned to 8 in our experiments.

%\vspace{-0.25cm}
\subsection{Implementation and Training Details}
%\vspace{-0.15cm}
\label{SEC3.4}
Our model was developed using PyTorch, and the batch size was set to 3 for the training on a single NVIDIA GeForce Titan-RTX GPU. 
We pre-trained a $ DEnet $ on Sceneflow dataset for 10 epochs, and utilized it as $ DEnet_a $ and $ DEnet_b $, except that the weights of the last two layers are initialized differently based on two random seeds. All weights of $ Confnet_a $ and $ Confnet_b $ were initialized differently.

We adopted an Adam optimizer ($ \beta_{1} $=0.9, $ \beta_{2} $=0.999) \cite{kingma2014adam} to train our model. $ \mathcal{L}_{self} $ and $ \mathcal{L}_{full} $ are calculated using the unlabeled and labeled samples, respectively. 
The gradients derived from minimizing $ \mathcal{L}_{full} $ update the weights of both disparity estimation and confidence networks, and those from minimizing $ \mathcal{L}_{self} $ only update the disparity estimation networks.
The losses on the reflective regions were masked out, and the reflective pixels were classified if their saturation value is less than 0.1 and intensity value is greater than 0.9.

In the training phase, we used a warm-up strategy.
Specifically, we first trained the two branches separately using labeled data, and then added unlabeled data for the following training using ACS and APS.
The number of warm-up and following semi-supervised training epochs varied according to different datasets, and is detailed in Sec.~\ref{sec:results}.
The learning rates in both warm-up and semi-supervised stages were set to 0.001, and were halved every quarter of the total epochs.
We randomly cropped image patch pairs with size of 256$ \times $256, and each patch pair was augmented by a random horizontal flip, gamma and brightness shifts. We set the maximum disparity value as $ S =192 $  in the calculation of cost volume. In the inference phase, the two branches of our method predict independently and we only report the prediction whose confidence map has a larger average value.

%\vspace{-0.25cm}
\section{Experimental Settings}
%\vspace{-0.15cm}
\subsection{State-of-the-arts methods}
We compare our method with several state-of-the-art methods including two traditional algorithms, six fully-supervised and five semi-supervised methods. 

The traditional algorithms for comparison are SGM \cite{hirschmuller2007stereo} and Stereo-UCL \cite{stoyanov2010real}, which are available in the OpenCV toolbox. In our experiments, SGM gets its best accuracy by setting aggregation paths Dir = 8, smoothness penalties P1 = 7 and P2 = 100. 
Stereo-UCL has no parameters to tune.

The fully-supervised methods for comparison include GANet~\cite{zhang2019ga}, PSMNet~\cite{chang2018pyramid}, HSMNet~\cite{yang2019hierarchical}, CDN~\cite{garg2020wasserstein}, CFNet~\cite{shen2021cfnet} and ACFNet~\cite{zhang2020adaptive}. 
These methods are open source, and we utilize the synthetic dataset (i.e., Sceneflow) to pretrain their models. 
When doing the fine-tuning for each method, we utilize the recommended parameters including the reported settings of learning rate, batch size, input size and optimizer.
%These methods all released their source codes, recommended training parameters and learned models. In our experiments, we use their released models for fine-tuning.
The semi-supervised methods for comparison include Smolyanskiy \textit{et al.}~\cite{smolyanskiy2018importance}, Ji \textit{et al.}~\cite{ji2019semi}, Tonioni \textit{et al.}~\cite{tonioni2019unsupervised}, Soft MT \cite{xu2021end} and Improved TSN \cite{shi2021semi}. The methods~\cite{smolyanskiy2018importance,ji2019semi,tonioni2019unsupervised} were originally proposed for depth estimation. We re-implement these three methods following their papers due to the lack of released source codes.
Soft MT \cite{xu2021end} is a competitive semi-supervised method for object detection. We compare with Soft MT since it belongs to the family of TSN like ours and released its source code. In Soft MT, the knowledge flow between the teacher and student models is still unidirectional, but the two models are trained synchronously in an exponential mean average (EMA) manner.
Improved TSN \cite{shi2021semi} is our previous work.

Note that, we replace the backbone with our employed $ DEnet $ in the two methods~\cite{ji2019semi,xu2021end}, because their developed backbones are for the task of monocular depth estimation and object detection respectively, and thus incompatible with the task we focus on in this paper.

\vspace{-0.3cm}
\subsection{Datasets}
%\vspace{-0.15cm}
\label{dataset}
This study includes four public datasets, i.e., SCARED \cite{allan2021stereo}, SERV-CT \cite{edwards2022serv}, KITTI 2012 \cite{Geiger2012CVPR} and ETH3D~\cite{schops2017multi}.

\subsubsection{SCARED}
SCARED is one of the MICCAI 2019 challenges~\cite{allan2021stereo}, and provided a training set acquired from 7 pigs. Each pig corresponds to 5 videos recorded by a binocular camera from 5 different views of the abdominal anatomy scene.
The video resolution is $ 1024 \times 1280 $, and the first frame of each video was marked as a keyframe. The ground-truth (GT) depth maps of keyframes were obtained by structured light projection directly, and those of the following video frames were approximated indirectly by depth interpolation using the transformation matrix relative to the keyframe camera position. Details about this dataset can be found in \cite{allan2021stereo}.

%There are 7 sub-datasets obtained from 7 pigs used for training, each of which contains 5 keyframes. The keyframes consist of binocular keyframe images, sequence video frames with the size of $ 1024 \times 1280 $ and the corresponding ground-truth (GT) depth. The video frames depth is interpolated from the keyframe depth according to the transformation matrix relative to the keyframe camera position.

There are two inaccuracies in the training set~\cite{allan2021stereo}.
First, the $ 4^{th} $ and $ 5^{th} $ subjects have to be excluded for calibration parameter errors.
Second, except keyframes, the video frames and GT depth maps are misaligned temporally. 
Thus, we only use the keyframes of the rest 5 pigs and their corresponding GT depth maps in our experiments. This leaves us $5 \times 5=25$ labeled keyframes in total for evaluation. 
In addition, we extract $250$ frames from each subject as unlabeled samples.

% and 

Recently, the challenge also released the official test set acquired from two extra pigs. Likewise, each pig corresponds to 5 binocular videos, and the first frame of each video was marked as a keyframe. Note that, the $ 5^{th} $ video of each pig only contains a single frame, that is, the video is identical to the keyframe. The GT depth maps of all frames were also released.

\begin{table*}[ht]
	\centering
	\footnotesize
	\caption{Average error of each sample in the 5-fold cross-validation on SCARED. The lowest error is marked in bold.}
	\begin{tabularx}{\linewidth}{p{2.96cm}>{\centering\arraybackslash}p{1.8cm}*4{>{\centering\arraybackslash}X}}
		\thickhline
		Methods & Supervision & $>$1px (\%) & $>$2px (\%) & $>$3px (\%) &  Disparity MAE (px)  \\ %&  \multicolumn{1}{p{4.49em}}{Depth(mm)}
		\hline
		SGM~\cite{hirschmuller2007stereo} & - & 30.04 $\pm$ 11.52  & 8.67 $\pm$ 6.22  & 3.71 $\pm$ 3.13 & 1.25 $\pm$ 0.14 \\% & 
		Stereo-UCL~\cite{stoyanov2010real} & - & 32.98 $\pm$ 7.90 & 10.93 $\pm$ 5.47 & 4.72 $\pm$ 3.04 & 1.21 $\pm$ 0.21 \\%& 
		\hline
		%		DispNetC~\cite{mayer2016large} &  & & & & &  \\
		
		GANet~\cite{zhang2019ga} & Fully & 25.36 $\pm$ 11.16 & 7.04 $\pm$ 4.97 & 2.96 $\pm$ 2.46 & 1.01 $\pm$ 0.20\\%& 1.260 
		
		PSMNet~\cite{chang2018pyramid} & Fully &26.53 $\pm$ 10.16 & 7.25 $\pm$ 4.91& 2.88 $\pm$ 2.46 & 0.92 $\pm$ 0.12\\%& 1.187
		
		HSMNet~\cite{yang2019hierarchical} & Fully & 25.11 $\pm$ 9.71 & 6.56 $\pm$ 4.56& 2.57 $\pm$ 2.42& 0.88 $\pm$ 0.12\\%& 1.115 
		
		CDN~\cite{garg2020wasserstein} & Fully & 24.03 $\pm$ 9.29 & 6.22 $ \pm $ 4.00 & 2.48 $ \pm $ 2.77 & 0.86 $\pm$ 0.10 \\
		
		CFNet~\cite{shen2021cfnet} & Fully & 23.60 $\pm$ 10.80& 6.19 $\pm$ 4.68& 2.48 $\pm$ 2.46& 0.86 $\pm$ 0.13 \\%& 1.120
		
		ACFNet~\cite{zhang2020adaptive}  & Fully & 23.77 $\pm$ 10.71 & 6.15 $\pm$ 4.69 & 2.54 $\pm$ 2.62 & 0.85  $\pm$ 0.11 \\
		
		\hline
		Smolyanskiy et al.~\cite{smolyanskiy2018importance} & Semi & 24.10 $\pm$ 10.71 & 6.55 $\pm$ 4.90& 2.72 $\pm$ 2.66&  0.86 $\pm$ 0.14\\%& 
		
		Ji et al.~\cite{ji2019semi} & Semi & 23.65 $\pm$ 9.95& 6.00 $\pm$ 4.38 & 2.39 $\pm$ 2.29& 0.84 $\pm$ 0.12 \\% &
		
		Tonioni et al.~\cite{tonioni2019unsupervised}& Semi & 23.23 $\pm$ 10.22 & 5.84 $\pm$ 4.27& 2.30 $\pm$ 2.20& 0.83 $\pm$ 0.12\\ %& 
		
		Soft MT~\cite{xu2021end}& Semi & 23.11 $\pm$ 9.99 & 5.81 $\pm$ 4.17& 2.25 $\pm$ 2.26& 0.82 $\pm$ 0.11\\ %& 
		
		Improved TSN~\cite{shi2021semi}& Semi & 22.32 $\pm$ 9.66& 5.51 $\pm$ 3.72 & 2.04 $\pm$ 1.87& 0.77 $\pm$ 0.11 \\
		
		Ours & Semi & \textbf{20.31 $\pm$ 9.83} & \textbf{5.01 $\pm$ 3.87} & \textbf{1.96 $\pm$ 1.88} & \textbf{0.74 $\pm$ 0.11}  \\ %& 0.985
		\thickhline
	\end{tabularx}%
	\label{tab:scared}%
\end{table*}%

\subsubsection{SERV-CT} The SERV-CT dataset was collected from 2 \emph{ex-vivo} porcine cadavers abdomen and each pig corresponds to 8 binocular image pairs with the size of $ 576\times 720 $. Both GT depth and disparity maps were provided.

%On SERV-CT, we conduct a leave-one-out evaluation. In each training round, we additionally include the 25 labeled samples of SCARED for training both fully-supervised and semi-supervised methods, and the 1000 unlabeled samples of SCARED for training the semi-supervised methods as well as ours. The total number of training epochs is set to 100.

\subsubsection{KITTI 2012} KITTI 2012 is a real-world dataset in the outdoor scenario, and contains 194 training and 195 testing stereo image pairs with the size of $ 376 \times 1240 $.
The training image pairs are provided corresponding GT disparity maps captured by LiDAR.

\subsubsection{ETH3D} ETH3D is a small real-world grayscale dataset with both indoor and outdoor scenes. It contains 27 labeled stereo image pairs for training and 20 stereo pairs for testing.

\vspace{-0.25cm}
\subsection{Evaluation Metrics}
%\vspace{-0.15cm}
For SCARED, we use the provided stereo calibration parameters to convert the GT depth into the GT disparity maps.
The evaluation metrics are disparity mean absolute error (MAE) and percentages of 1-px, 2-px, 3-px disparity outliers. 
Outliers are defined as the pixels whose disparity error is greater than a threshold (i.e., one, two or three pixels). 
%In addition, we use depth MAE to be consistent with the SCARED challenge.
On the two extra test subjects of SCARED, we also use the official metric, i.e., depth MAE, for evaluation.

For SERV-CT, the evaluation metrics are percentage of n-px disparity outliers and root mean square error (RMSE) for depth and disparity.
For KITTI, we also utilize the percentage of 3-px disparity outliers for evaluation.
%For dataset-specific metrics in the benchmarks, we follow the corresponding official definitions.
For ETH3D, we use the percentage of 1-px and 4-px disparity outliers, MAE and RMSE for evaluation. 
To be consistent with the ETH3D online leaderboard, we use the official names Bad 1.0, Bad 4.0 and AvgErr to refer to the first three metrics, respectively.
The calculation of MAE and RMSE is formulated in Eq. (\ref{equ:MAE}) and Eq. (\ref{equ:RMSE}).

\begin{small}
	\begin{equation}
		{MAE} = \frac{1}{N}\sum_{x,y} \big|D(x,y)-\hat{D}(x,y) \big|
		\label{equ:MAE}
	\end{equation}
\end{small}

\vspace{-0.3cm}
%\begin{small}
	\begin{equation}
		\small
		RMSE = \sqrt{\frac{1}{N}\sum_{x,y} \Bigl(D(x,y)-\hat{D}(x,y) \Bigr)^{2}}
		\label{equ:RMSE}
	\end{equation}
%\end{small}
where $ N $ represents the total number of labeled pixels.

\begin{table*}[t]
	\centering
	\footnotesize
	\caption{The challenge leaderboard of SCARED on two test subjects. kn is Depth MAE on the n-th keyframe of each subject. Avg. is average Depth MAE on all video frames. The lowest error is marked in bold, and the secondary lowest error is marked with underline.}
	\begin{tabularx}{\linewidth}{p{1.8cm}CCCCCCCCCCCC}
		\thickhline
		\multirow{2}{*}{Methods} & \multicolumn{6}{c}{The 1st test subject}& \multicolumn{6}{c}{The 2nd test subject}\\
		\cline{2-7} \cline{8-13}
		&k1&k2&k3&k4&k5&\multicolumn{1}{p{2.5em}}{Avg.}&k1&k2&k3&k4&k5&Avg. \\
		\hline
		\multicolumn{1}{l}{J.C. Rosenthal} & 8.25 &3.36 &2.21 &\underline{2.03} &1.33 &\multicolumn{1}{p{2.5em}}{3.44}& 8.26 &2.29 &7.04 &2.22 &0.42 &4.05  \\
		\multicolumn{1}{l}{Trevor Zeffiro} & 7.91 &2.97 &\underline{1.71} &2.52 &2.91 &\multicolumn{1}{p{2.5em}}{3.60}& 5.39 &1.67 &4.34 &3.18 &2.79 &3.47  \\
		\multicolumn{1}{l}{Dimitris Psychogyios 1}  & 7.73 &\underline{2.07} &1.94 &2.63 &\underline{0.62} &\multicolumn{1}{p{2.5em}}{3.00}& 4.85 &1.23 &3.52 &1.95 & \underline{0.41} &2.39 \\
		\multicolumn{1}{l}{Dimitris Psychogyios 2}  & \textbf{7.41} & \textbf{2.03} &1.92 &2.75 &0.65 &\multicolumn{1}{p{2.5em}}{\underline{2.95}}& 4.78 &1.19 &\underline{3.34} &1.82 & \textbf{0.36} & 2.30 \\
		\multicolumn{1}{l}{Sebatian Schmid}  & \underline{7.61} &2.41 &1.84 &2.48 &0.99 &\multicolumn{1}{p{2.5em}}{3.07}& \underline{4.33} &\underline{1.10} &3.65 &\underline{1.69} &0.48 & \underline{2.25}  \\
		
		\multicolumn{1}{l}{Ours}  & 7.98 &2.24 & \textbf{1.67} & \textbf{2.02} & \textbf{0.54} & \multicolumn{1}{p{2.5em}}{\textbf{2.89}}& \textbf{4.28} & \textbf{1.04} & \textbf{3.30} & \textbf{1.68} & \underline{0.41} &\textbf{2.14}  \\
		\thickhline
	\end{tabularx}%
	
	\label{tab:MEAN Scared challenge}%
\end{table*}%

\begin{table*}[t]
	\centering
	\footnotesize
	\caption{Average error of each sample in the leave-one-out cross-validation on SERV-CT. The lowest error is marked in bold.}
	\begin{tabularx}{\linewidth}{p{3cm}CCCCCCC}
		\thickhline
		{\multirow{2}{*}{Methods}}& \multicolumn{1}{c}{\multirow{2}{*}{Supervision}} &\multicolumn{2}{c}{$>3$px (\%)} & \multicolumn{2}{c}{Depth RMSE (mm)} & \multicolumn{2}{c}{Disparity RMSE (px)} \\ %&  \multicolumn{1}{p{4.49em}}{Depth(mm)}
		\cline{3-8}
		%		Methods & &not included & included & not included & included & not included & included \\
		& & Noc & All & Noc & All & Noc & All \\
		\hline
		
		HSMNet~\cite{yang2019hierarchical} & Fully & 5.63 & 8.52 & 2.13 & 2.93 & 1.59 & 2.18 \\
		
		CDN~\cite{garg2020wasserstein} & Fully & 5.57 & 8.20 & 2.15 & 2.95 & 1.57 & 2.21 \\	
		
		ACFNet~\cite{zhang2020adaptive}  & Fully & 5.51 & 8.11 & 2.16 & 2.97 & 1.54 & 2.19 \\
%		& 5.81 & 7.91 & 2.27 & 3.17 &1.54 & 2.19 \\
		
		CFNet~\cite{shen2021cfnet} & Fully & 5.38 & 7.73 & 2.14 & 2.97 & 1.50 &2.20 \\%& 1.120
		
		\hline
		
		Tonioni et al.~\cite{tonioni2019unsupervised} & Semi & 5.20 & 8.22 & 2.18 & 2.98 & 1.49 & 2.15 \\
		
		Soft MT~\cite{xu2021end} & Semi & 5.08  & 8.01  &  2.12  & 2.96 & 1.46 & 2.19 \\%& 
		
		Improved TSN~\cite{shi2021semi}& Semi & 4.29 & 6.96 & 2.24 & 2.89 & 1.51 & 2.10 \\
		
		Ours & Semi & \textbf{3.99} & \textbf{6.32} & \textbf{1.99} & \textbf{2.72}  & \textbf{1.34} & \textbf{1.86} \\ %& 0.985
		\thickhline
	\end{tabularx}%
	\label{tab:servct}%
\end{table*}%

\vspace{-0.25cm}
\section{Results and Discussions}
\label{sec:results}
%\vspace{-0.15cm}
\subsection{Comparison with State-of-the-arts}

\subsubsection{Evaluation on SCARED (5-fold cross-validation)} We conduct a 5-fold cross-validation on SCARED, that is, 20 labeled keyframes plus 1000 unlabeled video frames for training and 5 keyframes for test in each round of cross-validation. No subject is cross-used in the training and test folds. 
The number of warm-up and semi-supervised training epochs is set to 300 and 100 respectively.

Table \ref{tab:scared} lists the comparison results between our method and the state-of-the-art methods. 
The results are the average error of the 25 samples.
A lower value in this table indicates a better accuracy for all metrics.
%Note that, the two branches of our method predict independently in the reference phase, and we only report the prediction whose confidence map has the larger average value.

From Table \ref{tab:scared}, we can have three key observations:

(i) The accuracy of the two traditional methods is similar, but both relatively unsatisfying compared to that of the learning-based methods, demonstrating a powerful reasoning capability of CNNs;

(ii) The accuracy of the fully-supervised methods ($ 3^{rd} $ to $ 8^{th} $ rows) is mostly lower than that of the semi-supervised methods ($ 9^{th} $ to $ 14^{th} $ rows), because the latter can utilize unlabeled samples, and enjoy extra knowledge for improving the accuracy;

(iii) Our method achieves a superior accuracy by decreasing disparity MAE by \textbf{12.94\%} compared to the fully-supervised model ACFNet, and by \textbf{9.76\%} comparing to the semi-supervised model Soft MT. 
Compared to the second best method Improved TSN, our method reduces disparity MAE by \textbf{3.90\%}, which mainly benefits from two ameliorations, that is, the two learners in our method mutually guide each other in an adaptive bidirectional learning, and $ Confnet $ and $ DEnet $ are trained jointly by the constraint of disparity probability distribution. 
%In contrast, $ DEnet $ is isolated with $ Confnet $ in Improved TSN, and thus the predicted confidence has no feedback channel to direct $ DEnet $’s learning.

\begin{figure}[t]
	\centering
	\includegraphics[width=0.98\linewidth]{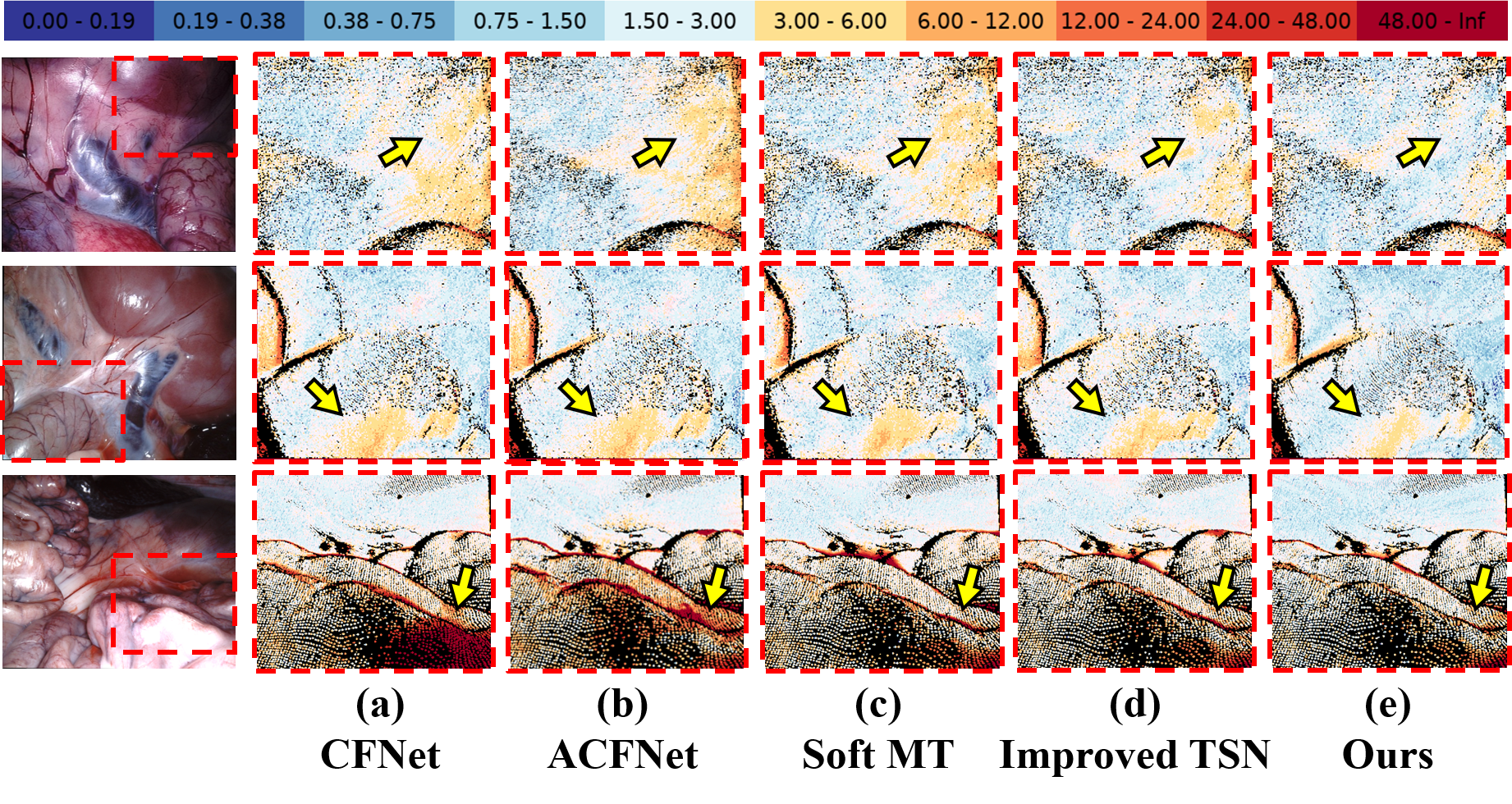}
	\caption{Error map of predicted disparities on three SCARED samples. The left-most col shows the input left images. (a)-(e) are results of the two best fully-supervised methods, the two best semi-supervised methods and our method. Different colors indicate different absolute distances between the GT and predicted disparity. The arrows in the $ 1^{st} $ and $ 2^{nd} $ rows show the errors at the flat areas, and the arrows in the $ 3^{rd} $ row shows the errors at the edge of the organ.}
	\vspace{-0.3cm}
	%		The red color indicates large difference, and the blue color means the difference is small. Areas with large errors are pointed by red dashed rectangles and zoomed in. 
	\label{fig:SCARED_Sota}
\end{figure}

Fig. \ref{fig:SCARED_Sota} visualizes the results of different methods, i.e., the two best fully-supervised methods, the two best semi-supervised methods and ours. 
The error map is obtained by calculating the absolute distance between the GT and predicted disparity for every position. 
As can be seen, the semi-supervised methods are better than the fully-supervised methods, and our method achieves the most accurate disparity prediction among them, especially at the flat areas (the $ 1^{st} $ and $ 2^{nd} $ rows) or the organ edges (the $ 3^{rd}$ row) indicated by the arrows.

\begin{figure}[t]
	\centering
	\includegraphics[width= \linewidth]{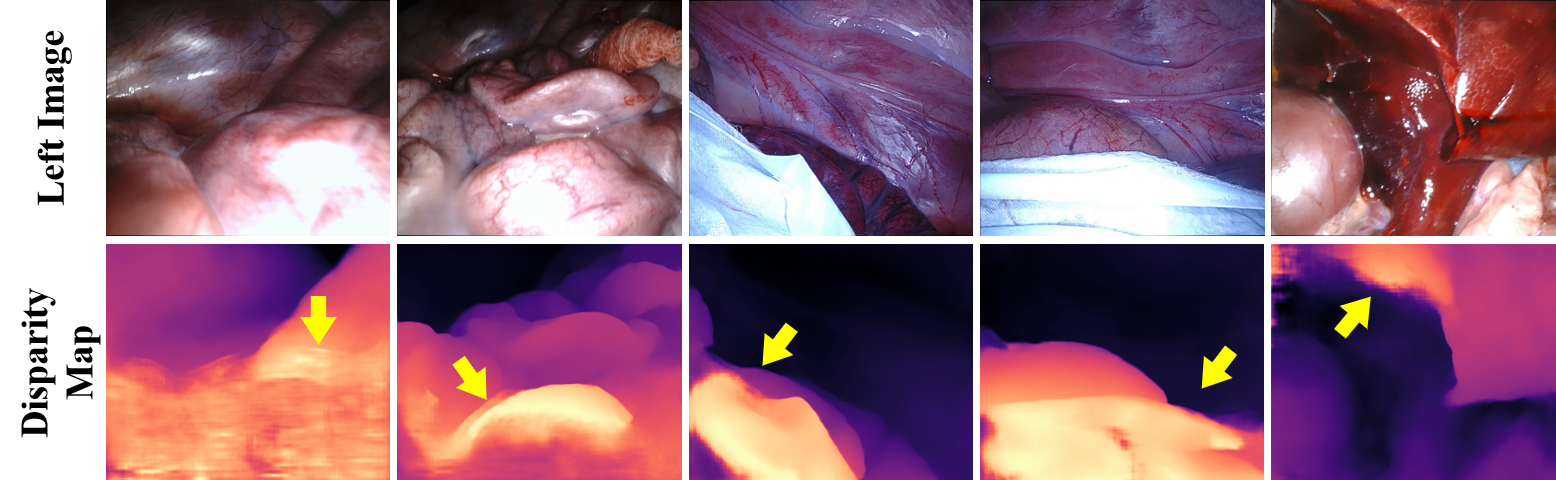}
	\caption{Some failure cases of our method on unlabeled SCARED samples. The yellow arrows indicate the challenging areas where our method failed to make accurate predictions. The surfaces in the $ 1^{st} $ and $ 2^{nd} $ column images have overexposed regions, the $ 3^{rd} $ and $ 4^{th} $ column images contain surgical bag which is nonexistent in the training data, and the $ 5^{th} $ column image has motion blur.}
	\vspace{-0.3cm}
	\label{fig:failure}
\end{figure}

Moreover, Fig.~\ref{fig:failure} illustrates several cases where our method fails to make accurate disparity predictions (indicated by arrows). The failure cases are mainly caused by three factors: (1) the severely-overexposed regions (see the $ 1^{st} $ and $ 2^{nd} $ columns), (2) nonexistent objects in the training data (see the surgical bag in the $ 3^{rd} $ and $ 4^{th} $ columns), and (3) the motion blur (see the $ 5^{th} $ column).

\subsubsection{Evaluation on SCARED (challenge leaderboard)} We use the labeled keyframes and unlabeled video frames of all 5 training subjects to re-train our model, and perform evaluation on the two official test subjects. 
Following the recommended evaluation protocol in \cite{allan2021stereo}, the evaluation is performed on all video frames except in which whose more than 90\% of GT maps is empty.
For each test subject, the depth MAE on each keyframe and the average depth MAE of all frames are reported. 
Table \ref{tab:MEAN Scared challenge} lists the comparison results between our method and other challenge participants whose results were reported in \cite{allan2021stereo}.

The first two rows of Table \ref{tab:MEAN Scared challenge} are the runner-up and winner of the challenge. After the challenge, the organizers allowed a period of late submission, and the three teams (the $ 3^{rd} $ to $ 5^{th} $ rows) were added into the leaderboard. 
As the official challenge winner, Trevor Zeffiro achieves an average depth MAE of 3.60 mm and 3.47 mm on the two test subjects, respectively. 
In comparison, our method reduces the average depth MAE by \textbf{19.72\%} and \textbf{38.33\%}. 

Among the late submissions, two teams achieve impressive results. Dimitris Psychogyios 2 finetuned HSMNet \cite{yang2019hierarchical} model on SCARED and achieves an average depth MAE of 2.95 mm and 2.30 mm. 
Sebatian Schmid used a 3D CV based method for stereo matching and achieves an average depth MAE of 3.07 mm and 2.25 mm.
Compared to these two teams, our method reduces the average depth MAE to 2.89 mm and 2.14 mm, which surpasses Dimitris Psychogyios 2 by \textbf{2.03\%} and \textbf{6.96\%}, and Sebatian Schmid by \textbf{5.86\%} and \textbf{4.89\%} on two test subjects. 

\subsubsection{Evaluation on SERV-CT (leave-one-out cross-validation)} We compare our method with the four best fully-supervised methods HSMNet, CDN, CFNet and ACFNet, and the three best semi-supervised methods Soft MT, Tonioni \textit{et al.} and Improved TSN, on SERV-CT. We conduct a leave-one-out cross-validation. In each round of cross-validation, we add the SCARED dataset to expand the training data. For the fully-supervised methods, 25 labeled images in SCARED plus 8 SERV-CT labeled images are used for training. For the semi-supervised methods, the above 33 labeled images plus the unlabeled video frame images selected from SCARED are used for training.
We set 1000 training epochs for fully-supervised methods. For the semi-supervised methods including ours, the number of warm-up epoch was 300, and the following semi-supervised training lasts for 100 epochs. All the initial learning rates are set to 0.001, and are halved every quarter of the epochs.

Table~\ref{tab:servct} lists the comparison results for both non-occluded (Noc) and all (All) pixels. The results are the average error of each sample (16 in total) in the leave-one-out cross-validation.
From these results, three conclusions can be made:

(i) The results of  Soft MT are only comparable to those of the fully-supervised methods. 
Therefore, the semi-supervised learning should not be overused for unlabeled images from a different data source;

(ii) Improved TSN successfully draws useful knowledge from the unlabeled samples, and thus achieves second-best accuracy in terms of most metrics. Therefore, error suppression on pseudo labels is of great importance when unlabeled data from other source are used;

(iii) Our method achieves a superior accuracy in terms of all metrics, and reduces disparity RMSE by at least \textbf{8.22\%} and \textbf{11.43\%} on non-occluded pixels and all pixels, respectively. 
%We believe the reason is that our proposed adaptive bidirectional learning makes the pseudo supervisions across two branches continually refined, and thus further maximizes the efficacy of unlabeled samples.

\begin{figure}[t]
	\centering
	\includegraphics[width=0.98\linewidth]{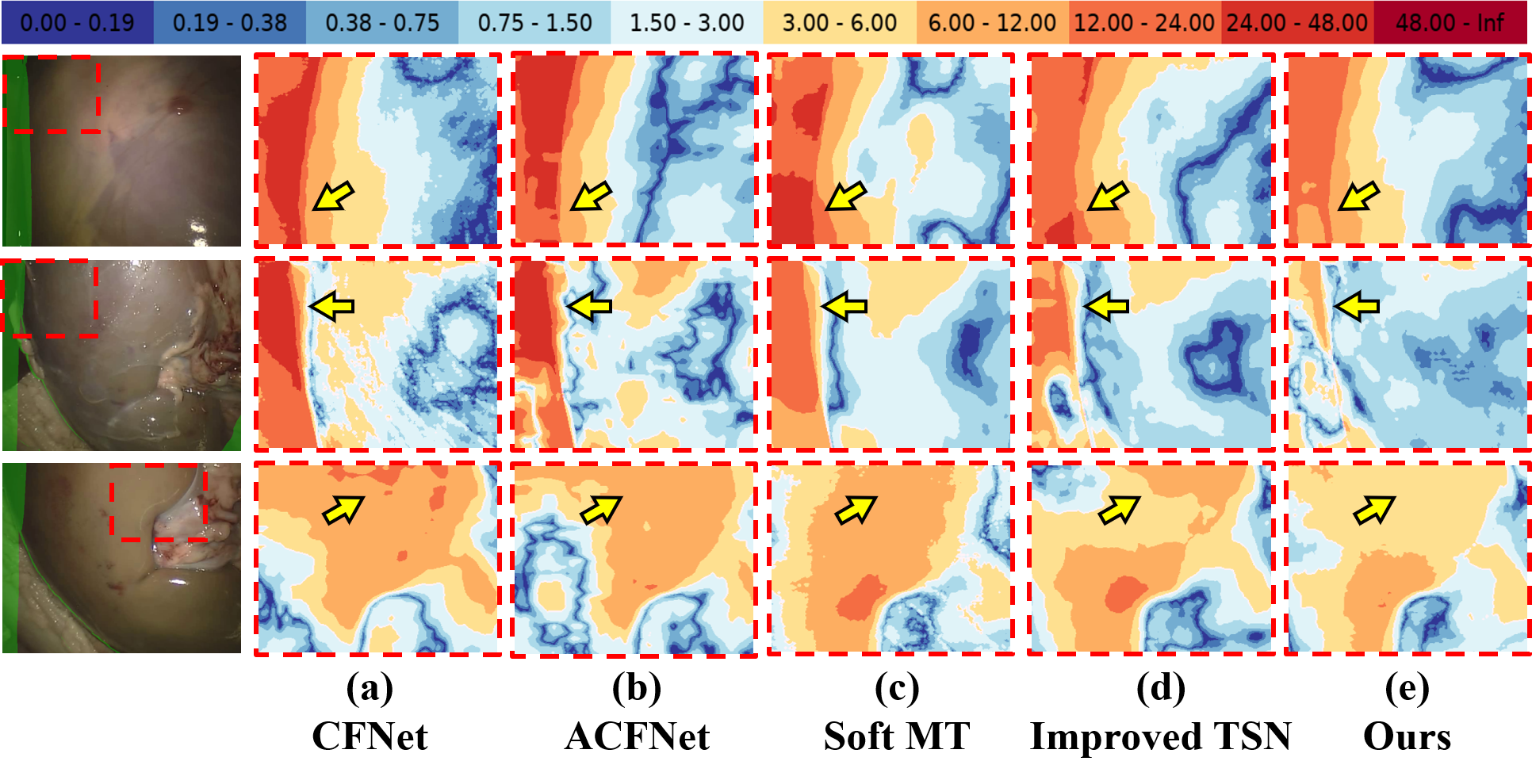}
	\caption{Error map of predicted disparities on three SERV-CT samples. The left-most col shows the input left images where the green transparent regions are occluded in the right counterpart. (a)-(e) are results of the two best fully-supervised methods, the two best semi-supervised methods and our method. Different colors indicate different absolute distances between the GT and predicted disparity.
	The arrows in the $ 1^{st} $ and $ 2^{nd} $ rows show the errors at low-textured organ surfaces, and the arrows in the $ 3^{rd} $ row shows the errors at the flat areas.}
	\vspace{-0.2cm}
	%		The red color indicates large difference, and the blue color means the difference is small. Areas with large errors are pointed by red dashed rectangles and zoomed in.
	\label{fig:SERVCT_SOTA}
\end{figure}

Fig.~\ref{fig:SERVCT_SOTA} visualizes error maps of the comparison methods.
The error maps of Improved TSN and ours are bluer in the occluded areas, which demonstrates the importance of error suppression of pseudo labels.
There are less red and no dark-red regions in our error maps, especially on the flat and low-textured organ surfaces indicated by the arrows.

\vspace{-0.4cm}
\subsection{Ablation Studies}

We conduct several ablation studies to verify the effectiveness of the two adaptive bidirectional supervisions, APS and ACS, and the joint learning of $ DEnet $ and $ Confnet $. SCARED and 5-fold cross validation are adopted.

\subsubsection{Effectiveness of APS and ACS} We train four variants of our dual-branch CNN by using APS and/or ACS and none of both. 
The model that is w/o both is treated as baseline, and identical to a single branch trained on only labeled samples.
The other three models are trained on both labeled and unlabeled samples.
The comparison results of the four variants are listed in Table \ref{tab:APS_ACS}. 
The statistical test is also performed on pairwise comparison of every two variants.
We used paired-samples T-test which is implemented in the SPSS software for statistical test .
%Since APS and ACS are designed for using unlabeled samples, the baseline model (see $ 1^{st} $ row) can only be trained on labeled samples, and the other three models are trained in a semi-supervised manner. 

Either APS or ACS can obtain an accuracy gain compared to baseline ($p=0.005$, $p=0.015$ in terms of Disparity MAE), but using ACS is more effective than using APS by reducing disparity MAE by 7.14\% ($p=0.06$).
% ($p$~=~1.51$e$-2). 
%This implies that only using APS might make the model learn features reflecting unreasonable matching relations, and thus lead to a right disparity value but incorrect probability distribution, while constraining the predicted distribution by ACS could force the model to make decisions by learning more reliable features. 
Also, APS and ACS are not mutually excluded, and together can further reduce disparity MAE by 7.50\% and 5.13\% ($p=0.002$, $p<0.001$), compared to APS only and ACS only, respectively.

It is worth noting that the model before using APS and ACS is inferior to the two TSN-based methods Soft MT and Improved TSN (Table \ref{tab:scared}), but becomes superior to all compared state-of-the-arts after being equipped with the adaptive bidirectional supervisions.

\begin{table}[t]
	\centering
	\small
	\caption{Comparison results of four variants of our method. The lowest error is marked in bold.}
	\begin{tabularx}{\linewidth}{p{2.8cm}>{\centering\arraybackslash}p{2cm}>{\centering\arraybackslash}X}
		\thickhline
		Variant Models & $>$3px (\%) & Disparity MAE (px) \\
		\hline
		%		\multicolumn{6}{c}{Disparity Estimation Results}\\
		%		\hline
		w/o Both (Baseline) & 2.44 $\pm$ 2.33& 0.84 $\pm$ 0.10\\ %& 0.985
		w/ APS only & 2.20 $\pm$ 2.16 & 0.80 $\pm$ 0.10\\ %0.815
		w/ ACS only & 2.09 $\pm$ 2.20& 0.78 $\pm$ 0.11\\
		w/ Both (Complete) & \textbf{1.96 $\pm$ 1.88} & \textbf{0.74 $\pm$ 0.11} \\
		\thickhline
	\end{tabularx}
	\label{tab:APS_ACS}
\end{table}

\begin{table}
	\centering
	\small
	\caption{
%		Comparison results between two variants using static or adaptive supervisions. SCU means the sum of average confidence of unlabeled samples.
Comparison results of four variants using Adaptive/Static Cross Supervision and Adaptive/Static Parallel Supervision. SCU means the sum of average confidence of unlabeled samples.}
	
	\begin{tabularx}{\linewidth}{p{1.8cm}>{\centering\arraybackslash}p{1.2cm}>{\centering\arraybackslash}p{1.6cm}>{\centering\arraybackslash}p{3cm}}
		\thickhline
		Supervision	& SCU & $>$3px (\%) & Disparity MAE (px) \\ %&  \multicolumn{1}{p{4.49em}}{Depth(mm)} 
		\hline
		%		Static & 903.6 & 2.10 $\pm$ 2.15& 0.77 $\pm$ 0.12 \\
		SCS $ \& $ SPS & 903.6 & 2.14 $\pm$ 2.22& 0.78 $\pm$ 0.11 \\
		SCS $ \& $ APS & 929.7 & 2.07 $ \pm $ 2.01 & 0.77 $\pm$ 0.12\\
		ACS $ \& $ SPS & 935.4 & 2.01 $ \pm $ 1.98 & 0.76 $\pm$ 0.11\\
		ACS $ \& $ APS & 968.7 & \textbf{1.96 $\pm$ 1.88} & \textbf{0.74 $\pm$ 0.11}\\
%		\hline
%		& & \multicolumn{2}{c}{p-values}\\
%		\hline
%		\multicolumn{2}{l}{Adaptive vs. Static} & 0.0424 & 0.0001 \\
		\thickhline
	\end{tabularx}%
	\label{tab:conf-K-tune}
\end{table}

\subsubsection{Adaptive vs. Static supervisions} 
%The confidence value controls the contribution of each branch’s pseudo supervisions provided to another by suppressing potential errors and softening labels of hard samples.
To further verify the effectiveness of both adaptive supervisions APS and ACS, we train four variants of Dual-Branch CNN, i.e., 'SCS \& SPS', 'SCS \& APS', 'ACS \& SPS' and 'ACS \& APS', on the SCARED dataset. Specifically, in the settings, Adaptive Parallel Supervision (APS) means the low-confident pseudo disparity values can be suppressed. Adaptive Cross Supervision (ACS) means the disparity distributions of hard samples can be softened by a changeable $\rho$. On the contrary, Static Parallel Supervision (SPS) means the pseudo disparity values are directly used as supervisions. Static Cross Supervision (SCS) means the disparity distributions of all samples are generated using a constant $\rho=1$.

Table \ref{tab:conf-K-tune} gives the comparison results of four variants using Adaptive/Static Cross Supervision and Adaptive/Static Parallel Supervision. By comparing the first three rows, both ACS and APS can obtain a significant error decrease with relative to the static version ($ p=0.009 $, $ p = 0.017 $ in terms of Disparity MAE). Also, ACS and APS are not mutually excluded, and combining them can further reduce disparity MAE by 3.90\% and 2.63\% by comparing the last row to the $ 2^{nd} $ and $ 3^{rd} $ rows ($ p = 0.025 $, $ p = 0.042 $ in terms of Disparity MAE). This well verifies the significant role of $ Confnet $ in both adaptive supervisions for our method.

We also calculate the sum of average confidence of unlabeled samples (SCU).
An increase of SCU indicates that more predictions can be effectively utilized as pseudo supervisions, reflecting a higher efficacy of unlabeled samples in the self-supervised training. 
From the $ 4^{th} $ column of Table \ref{tab:conf-K-tune}, the model with adaptive supervisions has a higher quality of pseudo supervisions. The complete version used the most unlabeled samples for training, resulting in a lower disparity error.

\subsubsection{Bidirectional vs. Unidirectional supervisions} 
%In this part, we train a variant of our dual-branch CNN, which utilizes predictions from one branch as pseudo supervisions to guide another, that is, the knowledge flow is unidirectional.
We compare the model with bidirectional supervision to that with unidirectional to verify the importance of the mutual guidance in the self-supervised learning. 
The comparison results are listed in Table \ref{tab:uni-bidirectional}.
The bidirectional knowledge flow can make the two branches eventually converge on more accurate disparity estimation by reducing Disparity MAE by 8.64\% compared to that with the unidirectional knowledge flow ($p<0.001$).
Also, SCU increases from 862.5 to 968.7 as the two branches are enabled to teach each other.

\begin{table}
	\centering
	\small
	\caption{Comparison results between two variants whose self-supervisions are bidirectional or unidirectional. SCU means the sum of average confidence of unlabeled samples.}
	\begin{tabularx}{\linewidth}{p{1.8cm}>{\centering\arraybackslash}p{1.2cm}>{\centering\arraybackslash}p{1.6cm}>{\centering\arraybackslash}p{3cm}}
		\thickhline
		Supervision & SCU & $>$3px (\%)& Disparity MAE (px)\\ 
		\hline
		%		\multicolumn{4}{c}{estimation results}\\
		%		\hline
		Unidirectional & 862.5 & 2.34 $\pm$ 2.30 & 0.81 $\pm$ 0.10\\
		Bidirectional & 968.7 & 1.96 $\pm$ 1.88 & 0.74 $\pm$ 0.11 \\
		\hline
		& & \multicolumn{2}{c}{p-values}\\
		\hline
		\multicolumn{2}{l}{Bidirc. vs. Unidirc.} & 0.0027 & 0.0008 \\
		\thickhline
	\end{tabularx}%
	\label{tab:uni-bidirectional}
\end{table}

\begin{figure}[t]
	\centering
	\includegraphics[width=0.98\linewidth]{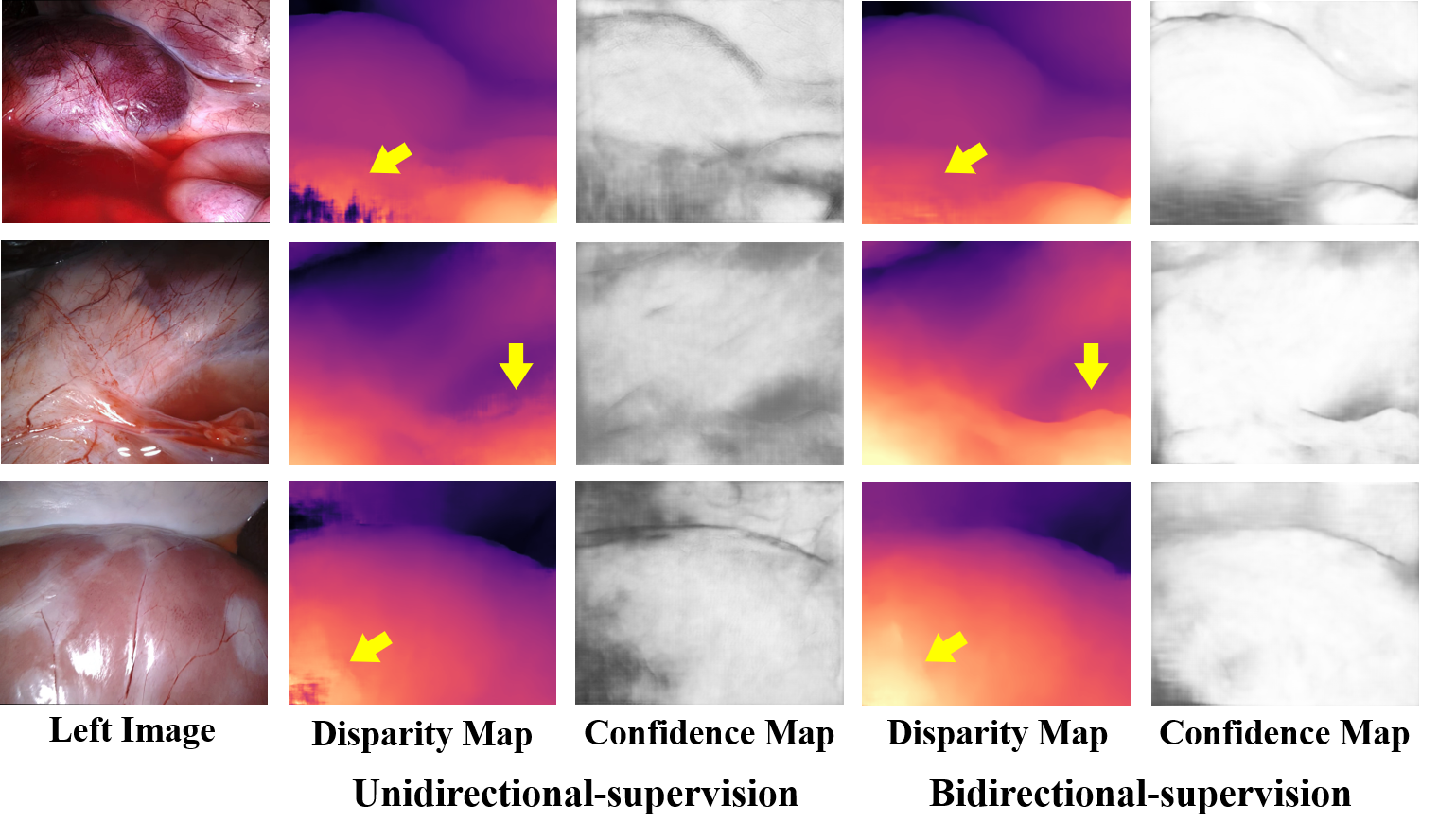}
	\caption{Three examples of the predicted disparity and confidence maps by the two variants of our method. Darker areas in the confidence map have lower confidence.}
	\vspace{-0.3cm}
	\label{fig:pseudochange}
\end{figure}

Fig. \ref{fig:pseudochange} further visualizes the disparity maps and confidence maps of unlabeled samples predicted by the comparison models. 
$ Confnet $ of both models successfully classifies the regions with possible errors as low confidence.
Overall, the confidence maps predicted under bidirectional supervisions are brighter (higher confidence) than those under unidirectional supervisions (the $3^{rd}$ vs. $5^{th}$ column).
Therefore, more unlabeled samples with high confidence can be utilized to improve the model accuracy, which is also evidenced by the increase of SCU in Table~\ref{tab:uni-bidirectional}.
As indicated by the yellow arrows in Fig.~\ref{fig:pseudochange}, the model trained under bidirectional supervisions predicts more reasonable disparities on the challenging regions, such as blood (the $ 1^{st}$ row), edge (the $ 2^{nd} $ row) and reflective surface (the $ 3^{rd} $ row), than that trained under unidirectional.

\subsubsection{Joint vs. Separate learning}  
We verify the accuracy gain by additionally introducing the distribution constraint $ \mathcal{L}_{dist} $ in the fully-supervised learning.
%During the fully-supervised training phase, our previous work (\cite{shi2021semi}) only optimized $ DEnet $ and $ Confnet $ separately. 
%In this paper, we employ an auxiliary loss $ \mathcal{L}_{dist} $ to constrain disparity probability distribution, forcing $ DEnet $ and $ Confnet $ to be trained jointly.
Therefore, we compare two variants of single-branch model with and without optimizing the disparity distribution constraint.
$DEnet$ and $Confnet$ are trained jointly if using the constraint, and are trained separately otherwise.

\begin{table}[t]
	\centering
	\small
	\caption{Comparison results between two variants of the single-branch CNN whose $DEnet$ and $Confnet$ are trained jointly or separately.}
	\begin{tabularx}{\linewidth}{p{2.5cm}>{\centering\arraybackslash}p{2.5cm}>{\centering\arraybackslash}p{2.8cm}}
		\thickhline
		Learning & $>$3px (\%) & Disparity MAE (px)	\\ %&  \multicolumn{1}{p{4.49em}}{Depth(mm)}  (\%) $ \downarrow $
		\hline
		Separate & 2.58 $\pm$ 2.34 & 0.87 $\pm$ 0.11 \\
		Joint & 2.44 $\pm$ 2.33 & 0.84 $\pm$ 0.10 \\
		\hline
		& \multicolumn{2}{c}{p-values}\\
		\hline
		Joint vs. Separate & 0.0452 & 0.0060 \\
		\thickhline
	\end{tabularx}%
	\label{tab:L-dist}
\end{table}

The comparison results are listed in Table \ref{tab:L-dist}.
Joint training achieves a significant improvement in terms of both metrics ($p<0.05$).
We believe the reason is that the confidence from $Confnet$ plays a role of softening labels of hard samples to enhance the learning of $ DEnet $ effectively.

\subsection{Learning Efficiency of Semi-supervised Methods}

\begin{figure}[t]
	\centering
	\includegraphics[width=\linewidth]{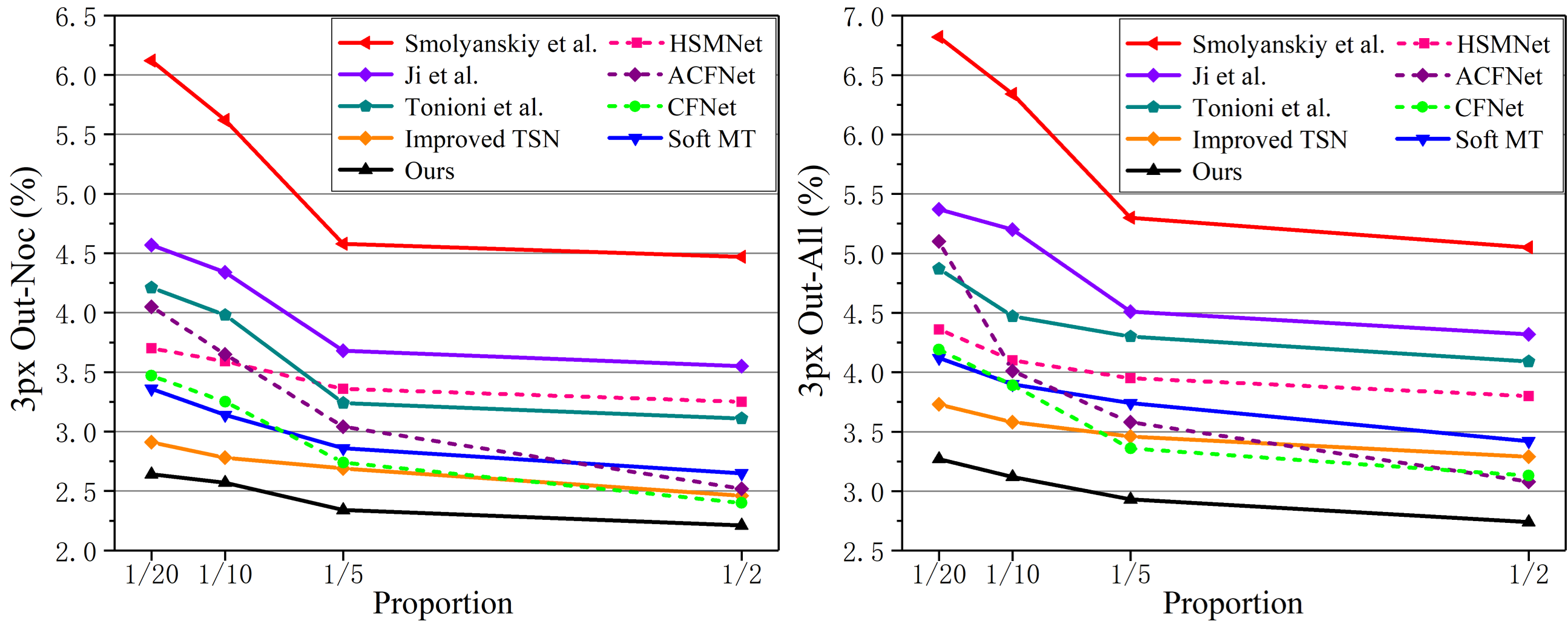}
	\caption{3px Out-Noc and Out-All error curves on KITTI of different methods which are trained under different proportions of available labels.}
	\vspace{-0.3cm}
	\label{fig:kitti}
\end{figure}

We use the KITTI 2012 dataset to evaluate training efficiency.
To this end, we split the dataset into 180 pairs for training and 14 for validation, and train each semi-supervised method four times.
In each time, we randomly selected a small proportion of training samples as labeled ones  (i.e., 5\%, 10\%, 20\%, and 50\%, respectively), and the rest as unlabeled ones, and the number of warm-up epoch and following semi-supervised training epochs is set to 100 and 600 respectively.
%the total number of training epochs is set to 600.
%we train four versions of each method and adjust the proportion of labels from 5\% to 50\% for training.  

Fig.~\ref{fig:kitti} illustrates the 3px Out-Noc and Out-All error on the validation set at different label proportions of the semi-supervised methods, i.e., Smolyanskiy \textit{et al.} \cite{smolyanskiy2018importance}, Ji \textit{et al.} \cite{ji2019semi}, Tonioni \textit{et al.}~\cite{tosi2019learning}, Soft MT \cite{xu2021end} and Improved TSN \cite{shi2021semi}, and our method. 
The errors of Soft MT, Improved TSN and ours are lower than others' by a large margin.
%, demonstrating the effectiveness of TSN-like knowledge distillation.
Moreover, our method consistently surpasses others regardless of the number of labels used for training. Especially, our method with only 1/20 labels achieves a comparable accuracy to Soft MT and Improved TSN with 1/5 or even 1/2 labels.

We also compare with the three best fully-supervised methods, i.e., HSMNet \cite{yang2019hierarchical}, ACFNet~\cite{zhang2020adaptive} and CFNet \cite{shen2021cfnet}.
When the proportion of labeled samples is less than 1/5, Improved TSN and our method are better than the three fully-supervised methods.
As the number of labels further increases, Improved TSN is inferior to ACFNet and CFNet, while our method is still the best.

\begin{table}[t]
	\centering
	\footnotesize
	\caption{Quantitative results on Kitti 2012 benchmark.}
	\begin{tabularx}{\linewidth}{p{1.8cm}CCCCCC}
		\thickhline
		{\multirow{2}{*}{Methods}}& \multicolumn{3}{c}{ 3px Out (\%)} & \multicolumn{3}{c}{ 5px Out (\%)} \\ %&  \multicolumn{1}{p{4.49em}}{Depth(mm)}
		\cline{2-7}
		%		Methods & &not included & included & not included & included & not included & included \\
		& Rank & Noc & All & Rank & Noc & All \\
		\hline
		
		\rule{0pt}{8pt}HSMNet~\cite{yang2019hierarchical} & 99 & 1.53 & 1.99 & 91 & 0.87 & 1.16 \\
		
		PSMNet~\cite{yang2019hierarchical} & 97 & 1.49 & 1.89 & 96 & 0.90 & 1.15 \\		
		
		GWCNet~\cite{guo2019group} & 67 & 1.32 & 1.70 & 66 & 0.80 & 1.03 \\
		
		CFNet~\cite{shen2021cfnet} & 51 & 1.23 & 1.58 & 41 & 0.74 & 0.94 \\
		
		GANet~\cite{zhang2019ga} & 42 & 1.19 & 1.60 & 49 & 0.76 & 1.02 \\
		
		ACFNet~\cite{zhang2020adaptive} & 37 & 1.17 & 1.54 & 56 & 0.77 & 1.01 \\
	
		Ours & \textbf{21} & \textbf{1.12} & \textbf{1.42} & \textbf{6} & \textbf{0.65} & \textbf{0.83} \\ %& 0.985
		\thickhline
	\end{tabularx}%
	\label{tab:kittibenchmark}%
\end{table}%

\begin{table}[t]
	\centering
	\footnotesize
	\caption{Quantitative results on ETH3D benchmark. The red superscript indicates the ranking of each method on the online leaderboard.}
	\begin{tabularx}{\linewidth}{p{1.8cm}CCCC}
		\thickhline
		Method & Bad 4.0 & Bad 1.0 & AvgErr & RMSE \\
		\hline

		\rule{0pt}{9pt}GWCNet~\cite{guo2019group} &0.50$ ^{\color{red}{171}} $&6.42$ ^{\color{red}{210}}$&0.35$ ^{\color{red}{200}}$&0.69$ ^{\color{red}{192}}$ \\
		
		HSMNet~\cite{yang2019hierarchical}&0.52$ ^{\color{red}{186}}$&4.20$ ^{\color{red}{172}}$&0.27$ ^{\color{red}{144}}$&0.67$ ^{\color{red}{183}}$\\
		
		GANet~\cite{zhang2019ga}&0.54$ ^{\color{red}{191}}$&6.56$ ^{\color{red}{206}}$&0.43$ ^{\color{red}{233}}$&0.75$ ^{\color{red}{211}}$\\
		
		PSMNet~\cite{yang2019hierarchical}&0.41$ ^{\color{red}{133}}$&5.02$ ^{\color{red}{192}}$&0.33$ ^{\color{red}{191}}$&0.66$ ^{\color{red}{181}}$\\
	
		CFNet~\cite{shen2021cfnet}&0.31$ ^{\color{red}{93}}$&3.31$ ^{\color{red}{123}}$&0.24$ ^{\color{red}{115}}$&0.51$ ^{\color{red}{98}}$\\
			
		Ours & \textbf{0.28}$ ^{\color{red}{84}}$ & \textbf{2.69}$ ^{\color{red}{81}}$ & \textbf{0.23}$ ^{\color{red}{102}}$ & \textbf{0.41}$ ^{\color{red}{27}}$ \\ 
		\thickhline
	\end{tabularx}%
	\label{tab:eth3dbenchmark}%
\end{table}%

\subsection{Results on two non-medical benchmarks}

Although our main focus is on medical images, we still benchmarked on the KITTI 2012 and ETH3D datasets to demonstrate the effectiveness of our method. 

To benchmark our method on KITTI 2012, we selected 180 labeled samples and 1,800 unlabeled samples for training, and the remaining 14 labeled samples are used as the validation set. We used the best model on the validation set to predict disparity maps for the 195 testing samples. We submitted the predicted disparity maps to the KITTI evaluation server for the evaluation. 
According to the online leaderboard\footnote{\url{https://www.cvlibs.net/datasets/kitti/eval_stereo_flow.php?benchmark=stereo}} at the time of submission, among all 223 submitted approaches, our method (named \emph{BSDual-CNN}) ranks $ 21^{th} $ in terms of 3px error, and $ 6^{th} $ in terms of 5px error. The benchmark results of the comparison methods included in this work and ours are listed in Table \ref{tab:kittibenchmark}. As can be seen, by effectively maximizing the usage of unlabeled data, our method significantly outperforms all the comparison methods on the KITTI official benchmark.

ETH3D provides 27 training samples with their GT disparity maps and 20 unlabeled test samples. We also introduced the 1,800 selected unlabeled images from KITTI for semi-supervised training since ETH3D didn't provide unlabeled samples. We submitted the results to the ETH3D online server for official evaluation.
According to the online leaderboard\footnote{\url{https://www.eth3d.net/low_res_two_view?metric=rms}}  at the time of submission, among all 300 submitted approaches, our method (named \emph{BSDual-CNN}) ranks $ 81^{th} $ in terms of Bad 1.0, and $ 27^{th} $ in terms of RMSE. The benchmark results of the comparison methods included in this work and ours are listed in Table \ref{tab:eth3dbenchmark}. It can be found that we achieved the best accuracy in the comparison methods, demonstrating the effectiveness of our method on small real-world datasets.

%\vspace{-0.1cm}
%\subsection{Bad cases}

\section{Conclusion}
In this paper, we successfully addressed the stagnant quality of pseudo labels caused by the teacher-to-student unidirectional knowledge flow in our previous work Improved TSN \cite{shi2021semi}.
Specifically, we proposed a novel dual-branch CNN with two kinds of adaptive bidirectional supervisions, named APS and ACS, in a semi-supervised manner.
With APS and ACS, the two branches can mutually guide each other by taking the opposite's predictions as pseudo supervisions.
Differently, APS constrains the predicted disparity values, while ACS forces the predicted disparity probability distributions to be unimodal.
This not only facilitates accurate disparity estimation, but also makes sure that the right disparity value is indeed from a reasonable distribution by learning explicit feature matching relations in the left-right image pair.
In addition, we introduced a confidence network in each branch to estimate the reliability of their predictions, and to adaptively refine pseudo supervisions flowing across branches. 
%Concretely, APS takes a higher possibility to suppress the pseudo supervision, and ACS adjusts the peak of target unimodal distribution to be lower and wider, if the pseudo supervision is with a lower confidence. 
Finally, with the well-tuned bidirectional supervisions, we maximized the efficacy of unlabeled samples, and jointly optimized the two branches to converge on a consistent and more accurate disparity prediction.

We conducted extensive and comprehensive comparisons with the state-of-the-arts on four public datasets, i.e., SCARED, SERV-CT, KITTI, ETH3D.
The comparison on SCARED demonstrated a superior accuracy of our method over other state-of-the-arts. 
The comparison on SERV-CT showed that the accuracy of our method can be more effectively enhanced by unlabeled samples from other data sources.
The comparison on KITTI validation set verified that our dual-branch CNN has a relatively higher efficiency of learning from unlabeled samples.
The results on KITTI and ETH3D benchmark respectively demonstrated the effectiveness of our method on common stereo matching datasets and real small datasets.
%In the future, we will extend our work to an under-determined task in endoscopic scenes, e.g., monocular depth estimation and multi-task, e.g., depth estimation and instrument segmentation jointly.
Four ablation studies were also conducted, and well demonstrated the effectiveness of our proposed two kinds of adaptive bidirectional supervisions, i.e., ACS and APS, and the joint training of $ DEnet $ and $ Confnet $ by additionally constraining the disparity probability distribution.

%\vspace{-0.3cm}
%\section*{Acknowledgments}

%\bibliography{refs}
\bibliographystyle{IEEEtran}
\bibliography{refs}

\begin{thebibliography}{10}
\providecommand{\url}[1]{#1}
\csname url@samestyle\endcsname
\providecommand{\newblock}{\relax}
\providecommand{\bibinfo}[2]{#2}
\providecommand{\BIBentrySTDinterwordspacing}{\spaceskip=0pt\relax}
\providecommand{\BIBentryALTinterwordstretchfactor}{4}
\providecommand{\BIBentryALTinterwordspacing}{\spaceskip=\fontdimen2\font plus
\BIBentryALTinterwordstretchfactor\fontdimen3\font minus
  \fontdimen4\font\relax}
\providecommand{\BIBforeignlanguage}[2]{{%
\expandafter\ifx\csname l@#1\endcsname\relax
\typeout{** WARNING: IEEEtran.bst: No hyphenation pattern has been}%
\typeout{** loaded for the language `#1'. Using the pattern for}%
\typeout{** the default language instead.}%
\else
\language=\csname l@#1\endcsname
\fi
#2}}
\providecommand{\BIBdecl}{\relax}
\BIBdecl

\bibitem{stoyanov2008intra}
D.~Stoyanov, G.~P. Mylonas, M.~Lerotic, A.~J. Chung, and G.-Z. Yang,
  ``Intra-operative visualizations: Perceptual fidelity and human factors,''
  \emph{Journal of Display Technology}, vol.~4, no.~4, pp. 491--501, 2008.

\bibitem{taylor2016medical}
R.~H. Taylor, A.~Menciassi, G.~Fichtinger, P.~Fiorini, and P.~Dario, ``Medical
  robotics and computer-integrated surgery,'' \emph{Springer handbook of
  robotics}, pp. 1657--1684, 2016.

\bibitem{maier2014comparative}
L.~Maier-Hein, A.~Groch, A.~Bartoli, S.~Bodenstedt, G.~Boissonnat, P.-L. Chang,
  N.~T. Clancy, D.~S. Elson, S.~Haase, E.~Heim \emph{et~al.}, ``Comparative
  validation of single-shot optical techniques for laparoscopic 3-d surface
  reconstruction,'' \emph{IEEE transactions on medical imaging}, vol.~33,
  no.~10, pp. 1913--1930, 2014.

\bibitem{mei2011building}
X.~Mei, X.~Sun, M.~Zhou, S.~Jiao, H.~Wang, and X.~Zhang, ``On building an
  accurate stereo matching system on graphics hardware,'' in \emph{2011 IEEE
  International Conference on Computer Vision Workshops (ICCV
  Workshops)}.\hskip 1em plus 0.5em minus 0.4em\relax IEEE, 2011, pp. 467--474.

\bibitem{zhang2019ga}
F.~Zhang, V.~Prisacariu, R.~Yang, and P.~H. Torr, ``Ga-net: Guided aggregation
  net for end-to-end stereo matching,'' in \emph{Proceedings of the IEEE/CVF
  Conference on Computer Vision and Pattern Recognition}, 2019, pp. 185--194.

\bibitem{shi2021semi}
H.~Shi, Z.~Wang, J.~Lv, Y.~Wang, P.~Zhang, F.~Zhu, and Q.~Li, ``Semi-supervised
  learning via improved teacher-student network for robust 3d reconstruction of
  stereo endoscopic image,'' in \emph{Proceedings of the 29th ACM International
  Conference on Multimedia}, 2021, pp. 4661--4669.

\bibitem{chang2013real}
P.-L. Chang, D.~Stoyanov, A.~J. Davison \emph{et~al.}, ``Real-time dense stereo
  reconstruction using convex optimisation with a cost-volume for image-guided
  robotic surgery,'' in \emph{International Conference on Medical Image
  Computing and Computer-Assisted Intervention}.\hskip 1em plus 0.5em minus
  0.4em\relax Springer, 2013, pp. 42--49.

\bibitem{penza2016dense}
V.~Penza, J.~Ortiz, L.~S. Mattos, A.~Forgione, and E.~De~Momi, ``Dense soft
  tissue 3d reconstruction refined with super-pixel segmentation for robotic
  abdominal surgery,'' \emph{International journal of computer assisted
  radiology and surgery}, vol.~11, no.~2, pp. 197--206, 2016.

\bibitem{rau2019implicit}
A.~Rau, P.~Edwards, O.~F. Ahmad, P.~Riordan, M.~Janatka, L.~B. Lovat, and
  D.~Stoyanov, ``Implicit domain adaptation with conditional generative
  adversarial networks for depth prediction in endoscopy,'' \emph{International
  journal of computer assisted radiology and surgery}, vol.~14, no.~7, pp.
  1167--1176, 2019.

\bibitem{allan2021stereo}
M.~Allan, J.~Mcleod, C.~C. Wang, J.~C. Rosenthal, K.~X. Fu, T.~Zeffiro, W.~Xia,
  Z.~Zhanshi, H.~Luo, X.~Zhang \emph{et~al.}, ``Stereo correspondence and
  reconstruction of endoscopic data challenge,'' \emph{arXiv preprint
  arXiv:2101.01133}, 2021.

\bibitem{mahmood2018deep}
F.~Mahmood and N.~J. Durr, ``Deep learning and conditional random fields-based
  depth estimation and topographical reconstruction from conventional
  endoscopy,'' \emph{Medical image analysis}, vol.~48, pp. 230--243, 2018.

\bibitem{luo2019details}
H.~Luo, Q.~Hu, and F.~Jia, ``Details preserved unsupervised depth estimation by
  fusing traditional stereo knowledge from laparoscopic images,''
  \emph{Healthcare technology letters}, vol.~6, no.~6, p. 154, 2019.

\bibitem{cui2019semi}
W.~Cui, Y.~Liu, Y.~Li, M.~Guo, Y.~Li, X.~Li, T.~Wang, X.~Zeng, and C.~Ye,
  ``Semi-supervised brain lesion segmentation with an adapted mean teacher
  model,'' in \emph{International Conference on Information Processing in
  Medical Imaging}.\hskip 1em plus 0.5em minus 0.4em\relax Springer, 2019, pp.
  554--565.

\bibitem{li2018semi}
X.~Li, L.~Yu, H.~Chen, C.-W. Fu, and P.-A. Heng, ``Semi-supervised skin lesion
  segmentation via transformation consistent self-ensembling model,''
  \emph{arXiv preprint arXiv:1808.03887}, 2018.

\bibitem{mayer2016large}
N.~Mayer, E.~Ilg, P.~Hausser, P.~Fischer, D.~Cremers, A.~Dosovitskiy, and
  T.~Brox, ``A large dataset to train convolutional networks for disparity,
  optical flow, and scene flow estimation,'' in \emph{Proceedings of the IEEE
  conference on computer vision and pattern recognition}, 2016, pp. 4040--4048.

\bibitem{godard2017unsupervised}
C.~Godard, O.~Mac~Aodha, and G.~J. Brostow, ``Unsupervised monocular depth
  estimation with left-right consistency,'' in \emph{Proceedings of the IEEE
  conference on computer vision and pattern recognition}, 2017, pp. 270--279.

\bibitem{ye2017self}
M.~Ye, E.~Johns, A.~Handa, L.~Zhang, P.~Pratt, and G.-Z. Yang,
  ``Self-supervised siamese learning on stereo image pairs for depth estimation
  in robotic surgery,'' \emph{arXiv preprint arXiv:1705.08260}, 2017.

\bibitem{huang2021self}
B.~Huang, J.-Q. Zheng, A.~Nguyen, D.~Tuch, K.~Vyas, S.~Giannarou, and D.~S.
  Elson, ``Self-supervised generative adversarial network for depth estimation
  in laparoscopic images,'' in \emph{International Conference on Medical Image
  Computing and Computer-Assisted Intervention}.\hskip 1em plus 0.5em minus
  0.4em\relax Springer, 2021, pp. 227--237.

\bibitem{smolyanskiy2018importance}
N.~Smolyanskiy, A.~Kamenev, and S.~Birchfield, ``On the importance of stereo
  for accurate depth estimation: An efficient semi-supervised deep neural
  network approach,'' in \emph{Proceedings of the IEEE conference on computer
  vision and pattern recognition workshops}, 2018, pp. 1007--1015.

\bibitem{laine2016temporal}
S.~Laine and T.~Aila, ``Temporal ensembling for semi-supervised learning,''
  \emph{arXiv preprint arXiv:1610.02242}, 2016.

\bibitem{tosi2019learning}
F.~Tosi, F.~Aleotti, M.~Poggi, and S.~Mattoccia, ``Learning monocular depth
  estimation infusing traditional stereo knowledge,'' in \emph{Proceedings of
  the IEEE/CVF Conference on Computer Vision and Pattern Recognition}, 2019,
  pp. 9799--9809.

\bibitem{yang2019hierarchical}
G.~Yang, J.~Manela, M.~Happold, and D.~Ramanan, ``Hierarchical deep stereo
  matching on high-resolution images,'' in \emph{Proceedings of the IEEE/CVF
  Conference on Computer Vision and Pattern Recognition}, 2019, pp. 5515--5524.

\bibitem{garg2020wasserstein}
D.~Garg, Y.~Wang, B.~Hariharan, M.~Campbell, K.~Q. Weinberger, and W.-L. Chao,
  ``Wasserstein distances for stereo disparity estimation,'' \emph{Advances in
  Neural Information Processing Systems}, vol.~33, pp. 22\,517--22\,529, 2020.

\bibitem{shen2021cfnet}
Z.~Shen, Y.~Dai, and Z.~Rao, ``Cfnet: Cascade and fused cost volume for robust
  stereo matching,'' in \emph{Proceedings of the IEEE/CVF Conference on
  Computer Vision and Pattern Recognition}, 2021, pp. 13\,906--13\,915.

\bibitem{zhang2020adaptive}
Y.~Zhang, Y.~Chen, X.~Bai, S.~Yu, K.~Yu, Z.~Li, and K.~Yang, ``Adaptive
  unimodal cost volume filtering for deep stereo matching,'' in
  \emph{Proceedings of the AAAI Conference on Artificial Intelligence},
  vol.~34, no.~07, 2020, pp. 12\,926--12\,934.

\bibitem{xu2021end}
M.~Xu, Z.~Zhang, H.~Hu, J.~Wang, L.~Wang, F.~Wei, X.~Bai, and Z.~Liu,
  ``End-to-end semi-supervised object detection with soft teacher,'' in
  \emph{Proceedings of the IEEE/CVF International Conference on Computer
  Vision}, 2021, pp. 3060--3069.

\bibitem{tonioni2019unsupervised}
A.~Tonioni, M.~Poggi, S.~Mattoccia, and L.~Di~Stefano, ``Unsupervised domain
  adaptation for depth prediction from images,'' \emph{IEEE transactions on
  pattern analysis and machine intelligence}, vol.~42, no.~10, pp. 2396--2409,
  2019.

\bibitem{Geiger2012CVPR}
A.~Geiger, P.~Lenz, and R.~Urtasun, ``Are we ready for autonomous driving? the
  kitti vision benchmark suite,'' in \emph{Conference on Computer Vision and
  Pattern Recognition (CVPR)}, 2012.

\bibitem{Menze2015CVPR}
M.~Menze and A.~Geiger, ``Object scene flow for autonomous vehicles,'' in
  \emph{Conference on Computer Vision and Pattern Recognition (CVPR)}, 2015.

\bibitem{kendall2017end}
A.~Kendall, H.~Martirosyan, S.~Dasgupta, P.~Henry, R.~Kennedy, A.~Bachrach, and
  A.~Bry, ``End-to-end learning of geometry and context for deep stereo
  regression,'' in \emph{Proceedings of the IEEE International Conference on
  Computer Vision}, 2017, pp. 66--75.

\bibitem{chang2018pyramid}
J.-R. Chang and Y.-S. Chen, ``Pyramid stereo matching network,'' in
  \emph{Proceedings of the IEEE Conference on Computer Vision and Pattern
  Recognition}, 2018, pp. 5410--5418.

\bibitem{guo2019group}
X.~Guo, K.~Yang, W.~Yang, X.~Wang, and H.~Li, ``Group-wise correlation stereo
  network,'' in \emph{Proceedings of the IEEE/CVF Conference on Computer Vision
  and Pattern Recognition}, 2019, pp. 3273--3282.

\bibitem{baek2022semi}
J.~Baek, G.~Kim, and S.~Kim, ``Semi-supervised learning with mutual
  distillation for monocular depth estimation,'' in \emph{2022 International
  Conference on Robotics and Automation (ICRA)}.\hskip 1em plus 0.5em minus
  0.4em\relax IEEE, 2022, pp. 4562--4569.

\bibitem{poggi2017quantitative}
M.~Poggi, F.~Tosi, and S.~Mattoccia, ``Quantitative evaluation of confidence
  measures in a machine learning world,'' in \emph{Proceedings of the IEEE
  International Conference on Computer Vision}, 2017, pp. 5228--5237.

\bibitem{hirschmuller2007stereo}
H.~Hirschmuller, ``Stereo processing by semiglobal matching and mutual
  information,'' \emph{IEEE Transactions on pattern analysis and machine
  intelligence}, vol.~30, no.~2, pp. 328--341, 2007.

\bibitem{wang2022rethinking}
J.~Wang and T.~Lukasiewicz, ``Rethinking bayesian deep learning methods for
  semi-supervised volumetric medical image segmentation,'' in \emph{Proceedings
  of the IEEE/CVF Conference on Computer Vision and Pattern Recognition}, 2022,
  pp. 182--190.

\bibitem{ronneberger2015u}
O.~Ronneberger, P.~Fischer, and T.~Brox, ``U-net: Convolutional networks for
  biomedical image segmentation,'' in \emph{International Conference on Medical
  image computing and computer-assisted intervention}.\hskip 1em plus 0.5em
  minus 0.4em\relax Springer, 2015, pp. 234--241.

\bibitem{muller2019does}
R.~M{\"u}ller, S.~Kornblith, and G.~E. Hinton, ``When does label smoothing
  help?'' \emph{Advances in neural information processing systems}, vol.~32,
  2019.

\bibitem{kingma2014adam}
D.~P. Kingma and J.~Ba, ``Adam: A method for stochastic optimization,''
  \emph{arXiv preprint arXiv:1412.6980}, 2014.

\bibitem{stoyanov2010real}
D.~Stoyanov, M.~V. Scarzanella, P.~Pratt, and G.-Z. Yang, ``Real-time stereo
  reconstruction in robotically assisted minimally invasive surgery,'' in
  \emph{International Conference on Medical Image Computing and
  Computer-Assisted Intervention}.\hskip 1em plus 0.5em minus 0.4em\relax
  Springer, 2010, pp. 275--282.

\bibitem{ji2019semi}
R.~Ji, K.~Li, Y.~Wang, X.~Sun, F.~Guo, X.~Guo, Y.~Wu, F.~Huang, and J.~Luo,
  ``Semi-supervised adversarial monocular depth estimation,'' \emph{IEEE
  transactions on pattern analysis and machine intelligence}, vol.~42, no.~10,
  pp. 2410--2422, 2019.

\bibitem{edwards2022serv}
P.~E. Edwards, D.~Psychogyios, S.~Speidel, L.~Maier-Hein, and D.~Stoyanov,
  ``Serv-ct: A disparity dataset from cone-beam ct for validation of endoscopic
  3d reconstruction,'' \emph{Medical image analysis}, vol.~76, p. 102302, 2022.

\bibitem{schops2017multi}
T.~Schops, J.~L. Schonberger, S.~Galliani, T.~Sattler, K.~Schindler,
  M.~Pollefeys, and A.~Geiger, ``A multi-view stereo benchmark with
  high-resolution images and multi-camera videos,'' in \emph{Proceedings of the
  IEEE Conference on Computer Vision and Pattern Recognition}, 2017, pp.
  3260--3269.

\end{thebibliography}

\end{document}